\documentclass[]{clv3}
\usepackage{microtype}
\usepackage{todonotes}

\usepackage{booktabs}
\usepackage{pgfplotstable}
\usepackage{pgfplots}

\pgfplotsset{compat=1.14}

\DeclareMathVersion{tabularbold}
\SetSymbolFont{operators}{tabularbold}{OT1}{cmr}{b}{n}

\usepackage{url}

\usepackage{tikz}
\usetikzlibrary{positioning, fit,  shapes.geometric, arrows.meta}
\tikzset{
	defaultfig/.style={
		node distance=1cm and 1.5cm,
		every text node part/.style= {
			align=center
		},
		word/.style= {
			blue,
			font=\itshape,
		},
		layer/.style= {
			rectangle, 
			black,
			draw
		},
		every node/.style ={anchor=base},
		every matrix/.style={ampersand replacement=\&, rounded corners=5pt, draw, column sep=15, row sep=20},
		refnode/.style={font={\footnotesize}, inner sep=0,outer sep=0, red, xshift=0.6cm},
	},
}

\newcommand{\refnode}[3][]{
	\node(#3ref)[#1,refnode]{Sect.\\ \ref{#2}};
	\draw[red] (#3ref.south west) to (#3);
}
\newcommand{\refnodeAR}[3][]{\refnode[above right=0.6 of #3,#1]{#2}{#3}}
\newcommand{\refnodeBR}[3][]{\refnode[below right=0.6 of #3,#1]{#2}{#3}}

\makeatletter
\g@addto@macro\@floatboxreset{\centering}
\makeatother

\usepackage{graphicx}
\graphicspath{{./figs/}, {./}}
\usepackage[space]{grffile}

\usepackage[subpreambles=false,group=false]{standalone}

\usepackage{amssymb}
\usepackage{amsmath}
\usepackage{mathtools}

\usepackage{cleveref}

\usepackage{url}
\newcommand{\parencite}{\citep}
\newcommand{\textcite}{\citet}

\newcommand{\natlang}[1]{\texttt{#1}}
\newcommand{\datapairs}{$\langle\text{color-name},\,(h,s,v)\rangle$}

\newcommand{\empmodel}{non-compositional baseline} 
\runningtitle{Learning of Colors from Color Names: Distribution and Point Estimation}
\runningauthor{Lyndon White et. al.}

\begin{document}
\title{Learning of Colors from Color Names: \\ Distribution and Point Estimation}
\author{Lyndon White\thanks{E-mail: {lyndon.white@invenialabs.co.uk}}}
\affil{Invenia Labs, Cambridge, UK\\ \& The University of Western Australia}
\author{Roberto Togneri\thanks{Email: {roberto.togneri@uwa.edu.au}}}
\affil{The University of Western Australia}
\author{Wei Liu\thanks{Email: {wei.liu@uwa.edu.au}}}
\affil{The University of Western Australia}
\author{Mohammed Bennamoun\thanks{Email: {mohammed.bennamoun@uwa.edu.au}}}
\affil{The University of Western Australia
}

\maketitle
\begin{abstract}
Color names are often made up of multiple words.
As a task in natural language understanding we investigate in depth the capacity of neural networks based on sums of word embeddings (SOWE), recurrence (LSTM and GRU based RNNs) and convolution (CNN), to estimate colors from sequences of terms.
We consider both point and distribution estimates of color.
We argue that the latter has a particular value as there is no clear agreement between people as to what a particular color describes -- different people have a different idea of what it means to be ``very dark orange'', for example.
Surprisingly, despite it's simplicity, the sum of word embeddings generally performs the best on almost all evaluations.
\end{abstract}

\section{Introduction}\label{sec:intro}

Consider that the word \texttt{tan} may mean one of many colors for different people in different circumstances: ranging from the bronze of a tanned sunbather, to the brown of tanned leather;
\texttt{green} may mean anything from \texttt{aquamarine} to \texttt{forest green};
and even \texttt{forest green} may mean the rich shades of a rain-forest, or the near grey of Australian bush land.
Thus the \emph{color intended} cannot be uniquely inferred from a color name. Without further context, it does nevertheless remain possible to estimate likelihoods of which colors are intended based on the population's use of the words.

Color understanding, that is, generating color from text, is an important subtask in natural language understanding.
For example, in a natural language enabled human-machine interface, when asked to select the \texttt{dark bluish green} object, it would be much useful if we could rank each object based on how likely its color matches against a learned distribution of the color name \texttt{dark bluish green}. 
This way if the most-likely object is eliminated (via another factor), the second most likely one can be considered.
A threshold can be set to terminate the search.
This kind of likelihood-based approach is not possible when we have only exact semantics based on point estimates.

Color understanding is a challenging domain, due to high levels of ambiguity, the multiple roles taken by the same words, the many modifiers, and the many shades of meaning.
In many ways it is a grounded microcosm of natural language understanding.
Due to its difficulty, texts containing color descriptions such as \texttt{the flower has petals that are bright pinkish purple with white stigma} are used to demonstrate the  capability of the-state-of-the-art image generation systems \parencite{reed2016generative, 2015arXiv151102793M}.
The core focus of the work we present is to map from the short-phrase descriptions of a color, to representation in a color-space such as HSV \parencite{smith1978color}.
The HSV color space is a grounded meaning space for the short phrase.
Due to this grounding, and the aforementioned linguistic phenomena, this is a particularly interesting short phrase understanding task.
Issues of illumination and perceived color based on context are considered out of the scope of this  article.

\subsection{Distribution vs. Point Estimation}
As illustrated, proper understanding of color names requires considering \emph{the color intended} as a random variable.
In other words, a color name should map to a distribution, not just a single point or region.
For a given color name, any number of points in the color-space could be intended, with some being more or less likely than others.
Or equivalently, up to interpretation, it may intend a region but the likelihood of what points are covered is variable and uncertain.
This distribution is often multimodal and has a high and asymmetrical variance, which further renders regression to a single point unsuitable.

A single point estimate does not capture the diverse nature of the color names adequately. Moreover, it is impossible to find the single best point estimation method.
For example: for a bimodal distribution, using the distribution's mean as a point estimate will minimize the total squared error,  but it will select a point in the valley between the peaks, which is less likely and less meaningful as a characterisation of that color.
Similarly for an asymmetrical distribution, where the mean will be off to one side of the peak.
Conversely, using the modes of the distribution the highest (most likely) peaks will be selected, but will on average be more incorrect as measured by the mean squared error.
The correct trade-off, if a point estimate is required, depends on the final use of the system.
Another problem is that point estimates do not capture the sensitivity.
In an asymmetrical distribution, having a point slightly off-centre in one direction may result in a very different probability,
this more generally holds for a narrow variance distribution.
Conversely for a very wide variance distribution (for example one approaching the uniform distribution) the point estimate value may matter very little with all points providing similar probabilities.
Color distributions are almost always multimodal or asymmetrical, and exhibit widely differing variances for different names.
This can be seen in the histograms of the training data shown in \Cref{fig:distout1} in \Cref{sec:qualitative-results}.
Note that while only a small (but particularly interesting) set of colors demonstrate multimodality in the hue channel, as noted by \textcite{mcmahan2015bayesian}, when considering all channels the other problematic features  abound. Asymmetry in-particular is ubiquitous in the value and saturation channels.
Given these issues, producing a point estimate has only limited value: estimating a distribution is a  more general task.
However we do consider the point estimation task, as it allows contrast in assessing the input module (SOWE/CNN/GRU/LSTM) of our proposed methods across the two different output modules (distribution/point estimation).

Generation of color from text has not received much attention in prior work.
To the best of our knowledge, the only similar work is \textcite{DBLP:journals/corr/KawakamiDRS16};
which only considers point estimation,
and uses a dataset containing far too few observations to allow for learning probability distributions from population usages of the color names.
\textcite{DBLP:journals/corr/KawakamiDRS16} uses a character sequence based model, rather than a word sequence model, which is inline with the very small amount of training data for each color name they have.
To our knowledge the generation of probability distributions in a color-space, from color names as considered as  sequences of words, has not been investigated at all by any prior work.
This paper is the first investigation of such kind.
There have been several works on the reverse problem \parencite{mcmahan2015bayesian,meomcmahanstone:color,2016arXiv160603821M}: the generation of a textual name for a color from a point in a color-space.
From these works on the reverse problem, there is a clear trend towards data-driven approaches in recent years where more color names and observations are used.
This motivates our own data-driven approach presented in this paper.


\subsection{Contributions}
{\bf Problem statement}: given a set of \datapairs{} pairs, we need to learn a mapping from any color-name, seen or unseen, to a color-value or a distribution in HSV color space.

We propose a neural network based architecture that can be broken down into an 
\textbf{input module}, which learns a vector representation of color-names,
 and a linked \textbf{output module}, which produces either a probability distribution or a  point estimate.
The \textbf{output module} uses a softmax output layer for probability distribution estimation,
or a novel HSV output layer for point estimation. 
To carry out the representational learning of color-names, four different color-name embedding learning models are investigated for use in the \textbf{input module}: Sum Of Word Embeddings (SOWE), Convolutional Neural Network (CNN) and two types of Recurrent Neural Network (LSTM and GRU RNNs).
All four input modules use pretrained FastText embeddings \parencite{bojanowski2016enriching} to represent the individual tokens making up the color names, but combine them using difference mechanisms.
The capacity of these input models is of primary interest to this work.

To evaluate and compare the three learning models, we designed a series of experiments to assess their capability in capturing compositionality of language used in color names.
These include:
\textbf{(1)} evaluation on all color names (full task);
\textbf{(2)} evaluation on  color names when the order of the words matters (order task);
\textbf{(3)} evaluation on color names which never occur in the training data in that exact form, but for which all terms occur in the training data (unseen combination task);
\textbf{(4)} qualitative demonstration of outputs for color names with terms which do not occur in the training data at all, but for which we know their word embeddings (embedding only task).

To express the estimated distribution for the output module, we discretize the HSV color-space to produce a histogram.
This allows us to take advantage of the well-known softmax based methods for the estimation of a probability mass distribution using a neural network.
An interesting challenge when considering this discretization is the smoothness of the estimate.
The true space is continuous, even if we are discretizing it at a resolution as high as the original color monitors used to collect the data.
Being continuous means that a small change in the point location in the color-space should correspond to a small change in how likely that point is according to the probability distribution.
Informally, this means the histograms should look smooth, and not spiky.
We investigated using a Kernel Density Estimation (KDE) based method for smoothing the training data, and further we conclude that the neural networks learn this smoothness.

We conclude that the simplest SOWE model is generally the best model for all tasks both for distribution and point estimation.
It is followed closely by the CNN; with the RNNs both performing significantly worse (see \Cref{sec:results}).
We believe that due to the nature of color understanding as a microcosm of natural language understanding, the results of our investigations have some implications for the capacity of how these models can be used for representing language compositionality in short phrase understanding.
%

\section{Related Work}\label{sec:related-work}
The understanding of color names has long been a concern of psycholinguistics and anthropologists \parencite{berlin1969basic,heider1972universals,HEIDER1972337,mylonas2015use}.
It is thus no surprise that there should be a corresponding field of research in natural language processing.

The earliest works revolve around explicit color dictionaries.
This includes the ISCC-NBS color system \parencite{kelly1955iscc} of 26 words, that are composed according to a context free grammar, such that phrases are mapped to single points in the color-space;
and the simpler, non-compositional, 11 basic colors of \textcite{berlin1969basic}.
Works including \textcite{Berk:1982:HFS:358589.358606,conway1992experimental,ele1994computational, mojsilovic2005computational, menegaz2007discrete,van2009learning}  propose methods for the automatic mapping of colors to and from these small manually defined sets of color names.
We note that \textcite{menegaz2007discrete,van2009learning} both propose systems that discretize the color-space, though to a much coarser level than we consider in this work.

The large Munroe dataset \parencite{Munroe2010XKCDdataset},
has allowed a data driven approach to natural language color problems.
In-contrast to earlier manually defined color dictionaries,
it has a large number of colors,  a non-trivial vocabulary,
and is sourced from a survey of hundreds of thousands of respondents.
Full details on this dataset can be found in \Cref{sec:full-task}.
The availability of a large color corpus has allowed machine learning based methods to be used in recent works including 
\textcite{mcmahan2015bayesian,meomcmahanstone:color,2016arXiv160603821M,acl2018WinnLighter} and this article.

\textcite{mcmahan2015bayesian} and \textcite{meomcmahanstone:color} present a Bayesian method for color estimation and color naming.
Their work primarily focuses on mapping from colors to to their exact names, the reverse of our task.
While their method is reversible: to go from exact color names to probabilities, they do not present any evaluations of this.
These works are based on defining fuzzy rectangular distributions in the color-space to cover the distribution estimated from the data, which are used in a Bayesian system to non-compositionally determine the color name.
This work focuses only on exact color names, where as later works consider the sequential nature of multi-word color names.

\textcite{2016arXiv160603821M} map a point in the color-space, to a sequence of probability estimates over color words.
They extend beyond all prior color naming systems to produce a compositional color namer based on the Munroe dataset.
Their method uses a recurrent neural network (RNN), which takes as input a color-space point, and the previous output word, and gives a probability of the next word to output -- this is a conditional language model.
We tackle the inverse problem, natural language understanding rather than generation.
Our distribution estimation models map from a sequence of terms, to a distribution in color-space.
Similarly, our point estimation models map from a sequence of terms to a single point in color-space.

\textcite{DBLP:journals/corr/KawakamiDRS16} proposes another compositional color naming model.
They use a per-character RNN and a variational autoencoder approach.
It is in principle very similar to \textcite{2016arXiv160603821M}, but functioning on a character, rather than word level.
The work by Kawakami et al. also includes a method for generating colors.
However they only consider the generation of point estimates, rather than distributions.
The primary focus of our work is on generating distributions.
The datasets used by Kawakami et al. contain only very small numbers of observations for each color name (often just one).
These datasets are thus not suitable for modeling the distribution in color-space as interpreted by a population.
Further, given the very small number of examples they are not well suited for use with word-based modeling: the character based modeling employed by Kawakami et al. is much more suitable.
As such, we do not attempt to compare with their work.

\textcite{DBLP:journals/corr/MonroeHGP17} present a neural network solution to a communication game, where a speaker is presented with three colors and asked to describe one of them, and the listener is to work out which color is being described.
Speaker and listener models are trained, using LSTM-based decoders and encoders, respectively.
The final time-step of their model produces a 100 dimensional representation of the description provided.
From this, a Gaussian distributed score function is calculated, over a high dimensional color-space defined by \textcite{2016arXiv160603821M}, which is then used to score each of the three options.
While this method does work with a probability distribution, as a step in its goal,
this distribution is always both symmetric and unimodal -- albeit in a high-dimensional color-space.

\textcite{acl2018WinnLighter} demonstrates a neural network for producing directional vectors in a color space indicating how comparatives such as \natlang{lighter} and \natlang{darker} change a color.
This effectively creates a ray (half-line) in color space along which possible colors described lie.
Their networks takes as its inputs a word embedding for a comparative adjective, and a point in RGB color-space.
It outputs a directional vector in the RGB space along which occurs the possible modified version of the input color point according to the given adjective.
The magnitude of this directional vector is trained such that adding it to the source color point, will give a good point estimate of the modified color.
For example mapping from \natlang{green} to a \natlang{darker green} is: $((164,227,77), \text{darker}) \mapsto (141, 190, 61)$ for a point estimate.
When using it for a ray estimate it is the half line from the first, through the second point, where every point further along the ray is \natlang{darker} than the earlier point.
The color adjectives may have up to two words, to allow for expressions such as \natlang{more neon}.
This is allowed by taking as a fixed sized input of two embeddings -- when only one input is required, the other is replaced by a zero vector.
Their training and evaluation is based on data sourced from the  Munroe dataset.

\section{Method}
\begin{figure}
	\resizebox{\textwidth}{!}{\begin{tikzpicture}[defaultfig]
		\node (data) [cylinder, shape border rotate=90, draw, minimum height=5 em] {Data};
		
		\node (prep) [matrix, right = of data] {
			\node{Training Data \\ Preparation}; \\
			\node(inputprep)[draw]{Input\\Tokenization};\\
			\node(outputprep)[draw]{Output\\Preparation}; \\
		};
		
		\refnodeAR[xshift=-0.4cm]{sec:input-data-preparation}{inputprep};
		\refnodeBR[xshift=-0.4cm]{sec:output-data-preparation}{outputprep};
		
		\node (input) [matrix, right = of prep] {
			\node{Input Modules}; \\
			\node(rnn1)[draw]{GRU RNN}; \\
			\node(rnn2)[draw]{LSTM RNN}; \\
			\node(sowe)[draw]{SOWE};\\
			\node(cnn)[draw]{CNN}; \\
		};
		\refnodeBR{sec:sowemod}{sowe};
		\refnodeBR{sec:cnnmod}{cnn};
		\refnodeAR[]{sec:rnnmod}{rnn1};
		\draw[red] (rnn1ref.south) to (rnn2.50);
		
		\node (output) [matrix, right = 1.4 of input, yshift=1em] {
			\node{Output Modules}; \\
			\node(softmax)[draw]{Softmax}; \\
			\node(hsvreg)[draw]{Novel HSV \\ Regression}; \\
		};
		\refnodeAR{sec:distmod}{softmax}
		\refnodeBR{sec:point-estimation}{hsvreg}

		\node(dist)[right=of softmax]{Probability\\Distribution\\Estimate};
		\node(point)[right=of hsvreg]{Point\\Estimate};
		
		\draw[-{Latex[length=2mm]}] (data) to (prep);
		\draw[-{Latex[length=2mm]}] (prep) to (input);
		\draw[-{Latex[length=2mm]}] (input) to (softmax);
		\draw[-{Latex[length=2mm]}] (input) to (hsvreg);
		\draw[-{Latex[length=2mm]}] (softmax) to (dist);
		\draw[-{Latex[length=2mm]}] (hsvreg) to (point);
		\end{tikzpicture}}
	\caption{The overall architecture of our system.\label{fig:arch}}
\end{figure}

\subsection{System Architecture}\label{sec:arch}
Our overall system architecture for all models is shown in \Cref{fig:arch}.
This shows how color names are transformed into distribution or point estimates over the HSV color-space.

\subsection{Input Data Preparation}\label{sec:input-data-preparation}
We desire a color prediction model which takes as input a sequence of words that make up the color's name rather than simply mapping from the whole phrase (whole phrase mapping does not scale to new user input, given the combinatorial nature of language).
Towards this end, color names are first tokenized into individual words.
For the input into our neural network based models, these words are represented with pretrained  word embeddings.

\subsubsection{Tokenization}
During tokenization a color name is split into terms with consistent spelling.
For example, \natlang{bluish kahki} would become the sequence of 3 tokens: [\natlang{blue}, \natlang{ish}, \natlang{khaki}].
Other than spelling, the tokenization results in the splitting of affixes and combining tokens (such as hyphens).
Combining tokens and related affixes affect how multiple colors can be combined.
The full list of tokenization rules can be found in the accompanying source code.
Some further examples showing how combining tokens and affixes are used and tokenized:
\begin{itemize}
	\item \natlang{blue purple} $\mapsto$ [\natlang{blue}, \natlang{purple}].
	\item \natlang{blue-purple} $\mapsto$ [\natlang{blue}, \natlang{-}, \natlang{purple}].
	\item \natlang{bluish purple} $\mapsto$ [\natlang{blue}, \natlang{ish}, \natlang{purple}]
	\item \natlang{bluy purple} $\mapsto$ [\natlang{blue}, \natlang{y}, \natlang{purple}]
	\item \natlang{blurple} $\mapsto$ [\natlang{blue-purple}]
\end{itemize}
The final example of \natlang{blurple} $\mapsto$ [\natlang{blue-purple}] is a special-case.
It is the only portmanteau in the dataset, and we do not have a clear way to tokenize it into a series of terms which occur in our pretrained embedding's vocabulary (see \Cref{sec:embeddings}).
The portmanteau \natlang{blurple} is not in common use in any training set used for creating word embeddings, so no pretrained embedding is available.%
\footnote{Methods do exist to generate embeddings for out of vocabulary words (like \natlang{blurple}), particularly with FastText embeddings \parencite{bojanowski2016enriching}. But we do not investigate those here.}
As such we handle it by treating it as the single token \natlang{blue-purple} for purposes of finding an embedding.
There are many similar hyphenated tokens in the pretrained embeddings vocabulary, however (with that exception) we do not use them as it reduces the sequential modeling task to the point of being uninteresting.

\subsubsection{Embeddings}\label{sec:embeddings} 
All our neural network based solutions incorporate an embedding layer.
This embedding layer maps from tokenized words to vectors.
We make use of 300 dimensional pretrained FastText embeddings \parencite{bojanowski2016enriching}\footnote{Available from \url{https://fasttext.cc/docs/en/english-vectors.html}}.

The embeddings are not trained during the task, but are kept fixed.
As per the universal approximation theorem \parencite{leshno1993uat, SONODA2017uat} the layers above the embedding layer allow for arbitrary continuous transformations.
By fixing the embeddings, and learning this transformation,
we can produce estimates of colors from embeddings alone, without any training data at all, as shown in \Cref{sec:embeddingonly}.

\subsection{Output Data Preparation for Training Distribution Estimation Models}\label{sec:output-data-preparation}
To train the distribution estimation models we need to preprocess the training data into a distribution.
The raw training data itself, is just a collection of 
 \datapairs{} pairs -- samples from the distributions for each named-color.
This is suitable for training the point estimation models, but not for the distribution estimation models .
We thus convert it into a tractable form, of one histogram per output channel -- by assuming the output channels are conditionally independent of each other.

\subsubsection{Conditional Independence Assumption} \label{sec:conditional-independence-assumption}
We make the assumption that given the name of the color, the distribution of the hue, saturation and value channels are independent.
That is to say, it is assumed if the color name is known, then  knowing the value of one channel would not provide any additional information as to the value of the other two channels.
The same assumption is made, though not remarked upon, in \textcite{meomcmahanstone:color} and \textcite{mcmahan2015bayesian}.
This assumption of conditional independence allows considerable saving in computational resources.
Approximating the 3D joint distribution as the product of three 1D distributions decreases the space complexity from $O(n^3)$ to $O(n)$ in the discretized step that follows.

Superficial checks were carried out on the accuracy of this assumption.
Spearman's correlation on the training data suggests that for over three quarters of all color names, there is only weak correlation between the channels for most colors (\mbox{Q3 = 0.187}).
However, this measure underestimates correlation for values which have a circular relative value, such as hue.
Of the 16 color-spaces evaluated, HSV had the lowest correlation by a large margin.
Full details, including the table of correlations, are available in supplementary materials (\Cref{sec:corrind}).
These results are suggestive, rather than solidly indicative, on the degree of correctness of the conditional independence assumption.
We consider the assumption sufficient for this investigation; as it does not impact on the correctness of results.
A method that does not make this assumption may perform better when evaluated using the same metrics we use here.

\subsubsection{Discretization} \label{sec:discretization}
For distribution estimation, our models are trained to output histograms.
This is a discretized representation of the continuous distribution.
Following standard practice, interpolation-based methods can be used to handle it as a continuous distribution.
By making use of the conditional independence assumption (see \Cref{sec:conditional-independence-assumption}), we output one 256-bin histogram per channel.
We note that 24-bit color (as was used in the survey that collected the dataset) can have all the information captured by a 256 bin discretization  per channel.
24 bit color allows for a total of $2^{24}$ colors to be represented, and even one-hot encoding for each of the 256 bin discretization channels allows for the same.
As such there is no meaningful loss of information during the discretization step when using histograms over a truly continuous estimation method, such as a Gaussian mixture model.
Although such models may have other advantages (such as the apriori information added by specifying the distribution), we do not investigate them here, instead considering the simplest non-parametric estimation model (the histogram), which has the simple implementation in a neural network using a softmax output layer.

Discretizing the data in this way is a useful solution used in several other machine learning systems.
\textcite{oord2016pixel, DBLP:journals/corr/OordDZSVGKSK16} apply a similar discretization step and found their method to outperform the more complex truly continuous distribution outputting systems.

For training purposes we thus convert all the observations into histograms.
One set of training histograms is produced per color description present in the dataset -- that is to say a training histogram is created for \natlang{brownish green}, \natlang{greenish brown}, \natlang{green} etc.
We perform uniform weight attribution of points to bins as described by \textcite{jones1984remark}.
In-short, this method of tabulation is to define the bins by their midpoints, and to allocate probability mass to each bin based on how close an observe point is to the adjacent midpoints.
A point precisely on a midpoint would have all its probability mass allocated to that bin;
whereas a point halfway between midpoints would have 50\% of its mass allocated to each.
For example were we to have bins with midpoints at 1 and 3:
then observation at 2, then 50\% of probability mass for this observation would be allocated to the bin with midpoint 1, and 50
Whereas if there observation was at 2.5 then 25\% of its mass would be allocated to the bin at 1, and 25\% of its mass to the bin at 3.
Our bins are at much finer resolution than this example, dividing the space between 0 and 1 into 256 bins.
This form of tabulation commonly used during the first step of performing kernel density estimation, prior to the application of the kernel.

\subsection{Color Name Representation Learning (Input Modules)}\label{sec:inputmod}
For each of the models investigated we define an input module.
This module handles the input of the color name, which is provided as a sequence of tokens.
It produces a fixed sized dense representation of the color name,
which is the input to the output module \Cref{sec:outputmod}).
In all models the input and output modules are trained concurrently as a single system.
%
%

\subsubsection{Recurrent Neural Networks (GRU and LSTM  RNNs)}\label{sec:rnnmod}
\begin{figure}
	\begin{tikzpicture}[defaultfig]
	\node (hiddenoutput)[layer] at (0,0) {ReLU};
	\node (output)[dotted, layer, above=1 of hiddenoutput] {Output Module};
	\draw[->] (hiddenoutput) to (output);
	
	\node (GRU1)[layer, below = of hiddenoutput]{GRU/\\LSTM};
	
	\foreach[count=\i from 1] \j in {2,...,5}
	{
		\node (GRU\j)[layer, left = of GRU\i]{GRU/\\LSTM};
		\draw[->] (GRU\j) to (GRU\i);
	}
	
	\foreach[count=\i from 1] \word in {$\langle EOS \rangle$, green, ish, blue, light}
	{
		\node (emb\i)[layer, below = of GRU\i]  {Embedding};
		\node (word\i)[word, below = of emb\i]{\word};
		\draw[->] (word\i) to  (emb\i);
		\draw[->] (emb\i) to (GRU\i);
		\node[draw,dashed,fit= (emb\i)  (word\i)  (GRU\i)] {};
	}

	\draw[->] (GRU1) to (hiddenoutput);
	\end{tikzpicture}
	
	\caption{The LSTM/GRU Input module
		for the example input \natlang{light greenish blue}. Each dashed box represents 1 time-step. \label{fig:rnnmod}.
	}
\end{figure}

A Recurrent Neural Network (RNN) is a common choice for this kind of task,
due to the variable length of the input.
We consider two "flavours" of RNN:  Gated Recurrent Unit (GRU) networks \parencite{cho2014properties}, and  Long Short Term Memory (LSTM) networks \parencite{hochreiter1997long,gers1999learning}.
They differ only in their recurrent unit's internal structure.
The general structure of both input modules is  shown in \Cref{fig:rnnmod}.
It is similar to \textcite{2016arXiv160603821M}, or indeed to most other word sequence learning models.
Each word is first transformed to an embedding representation.
This representation is trained with the rest of the network allowing per word information to be efficiently learned.
The embedding is used as the input for the recurrent unit, either a GRU or an LSTM depending on the model considered.
The stream is terminated with an End of Stream (\natlang{<EOS>}) pseudo-token,
represented using a zero-vector.
The output of the last time-step is fed to a Rectified Linear Unit (ReLU), and then to the output model.

During preliminary investigations we also considered a vanilla RNN (that is to say one without any gating). Early results on the development set suggested that it performed only marginally worse than the GRU or LSTM networks.
That it does not perform much worse than the models with features to improve memory is unsurprising, as the color names have at most 5 terms.
We constrained our full investigation to the more popular GRU and LSTM networks.

\subsubsection{Sum of Word Embeddings (SOWE)}\label{sec:sowemod}
Using a simple sum of word embeddings as a layer in a neural network is less typical than an RNN structure,
though it is well established as a useful representation, and has been used as an input to other classifiers such as support vector machines (e.g. as in \textcite{White2015SentVecMeaning,novelperspective}).
Any number of word embeddings can be added to the sum, thus allowing it to handle sequences of any length.
However, it has no representation of the order.
The structure we used is shown in \Cref{fig:sowemod}.

\begin{figure}
	\begin{tikzpicture}[defaultfig]
	
	\node (GRU1)[]{};
	
	\foreach[count=\i from 1] \j in {2,...,5}
	{
		\node (GRU\j)[left = 2 of GRU\i]{};
	}
	
	\node (sum)[layer, above= of GRU3, xshift=1cm]{$\sum$};
	
	\foreach[count=\i from 1] \word in {green, ish, blue, light}
	{
		\node (emb\i)[layer, below = of GRU\i]  {Embedding};
		\node (word\i)[word, below = of emb\i]{\word};
		\draw[->] (word\i) to  (emb\i);
		\draw[->] (emb\i) to (sum);
	}
	
	\node (hiddenoutput)[layer, above=of sum] {ReLU};
	\node (output)[dotted, layer, above=1 of hiddenoutput] {Output Module};
	\draw[->] (sum) to (hiddenoutput);
	\draw[->] (hiddenoutput) to (output);
	
	\end{tikzpicture}
	
	\caption{The SOWE input module for the example input \natlang{light bluish green}}
	\label{fig:sowemod}
\end{figure}

\subsubsection{Convolutional Neural Network(CNN)}\label{sec:cnnmod}
A convolutional neural network (shown in \Cref{fig:cnnmod}) can be applied to the task by applying 2D convolution over the stacked word embeddings.
We use 64 filters of size between one and five.
Five is number of tokens in the longest color-name,
so this allows it to learn full length relations.

\begin{figure}
	\begin{tikzpicture}[defaultfig]
	
	\node (GRU1)[]{};
	
	\foreach[count=\i from 1] \j in {2,...,5}
	{
		\node (GRU\j)[left = 2 of GRU\i]{};
	}
	
	\node (hstack)[layer, above= of GRU3, xshift=1cm]{stack into grid};
	
	\foreach[count=\i from 1] \word in {green, ish, blue, light}
	{
		\node (emb\i)[layer, below = of GRU\i]  {Embedding};
		\node (word\i)[word, below = of emb\i]{\word};
		\draw[->] (word\i) to  (emb\i);
		\draw[->] (emb\i) to (hstack);
	}
	
	\node (conv)[layer, above= 0.5 of hstack]{2D Convolution};
	\node (innerconv)[layer, above= 0.5 of conv]{ReLU};
	\node (pool)[layer, above= 0.5 of innerconv]{MaxPooling};
	\node (hiddenoutput)[layer, above= 0.5 of pool] {ReLU};
	\node (output)[dotted, layer, above=1 of hiddenoutput] {Output Module};
	\draw[->] (hstack) to (conv);
	\draw[->] (conv) to (innerconv);
	\draw[->] (innerconv) to (pool);
	\draw[->] (pool) to (hiddenoutput);
	\draw[->] (hiddenoutput) to (output);
	
	
	\end{tikzpicture}
	
	\caption{The CNN input module for the example input \natlang{light bluish green}.}
	\label{fig:cnnmod}
\end{figure}

\subsection{Distribution and Point Estimation (Output Modules)}\label{sec:outputmod}
On top of the input module, we define an output module to suit the neural network for the task of either distribution estimation or point estimation.
The input module defines how the terms are composed into the network.
The output module defines how the network takes its hidden representation and produces an output.

\subsubsection{Distribution Estimation}\label{sec:distmod}
The distributions are trained to produce the discretized representation as discussed in \Cref{sec:discretization}.
Making use of the conditional independence assumption (see \Cref{sec:conditional-independence-assumption}), we output the three discretized distributions.
As shown in \Cref{fig:distoutmod}, this is done using three softmax output layers -- one per channel.
They share a common input, but have separate weights and biases.
The loss function is given by the sum of the cross-entropy for each of the three softmax outputs.

\begin{figure}
	\newcommand{\picwidth}{60pt}
	\begin{tikzpicture}[defaultfig]
	
	\node (input)[layer, dotted]{Input Module};
	
	\node(outHue)[layer, above left = of input] {Softmax};
	\node(outSaturation)[layer, above = of input] {Softmax};
	\node(outValue)[layer, above right = of input] {Softmax};
	
	\foreach \p in {Hue, Saturation, Value} 
	{
		\draw[->] (input) to (out\p);
		
		\node(plot\p)[above = of out\p, text width=\picwidth]{
			\includegraphics[width=\picwidth]{netdia/\p}
			\\
			\p
		};
		\draw[->] (out\p) to (plot\p);
	}
	\end{tikzpicture}
	
	\caption{The Distribution Output Module \label{fig:distoutmod}}
\end{figure}

\subsubsection{Point Estimation}\label{sec:point-estimation}
Our point estimation output module is shown in \Cref{fig:pointoutmod}.
The hidden-layer from the top of the input module is fed to  four single output neurons.\footnote{Equivalently these four single neurons can be expressed as a layer with four outputs and two different activation functions.}
Two of these use the sigmoid activation function (range 0:1) to produce the outputs for the saturation and value channels.
The other two use the tanh activation function (range -1:1), and produce the intermediate output that we call $y_{shue}$ and $y_{chue}$ for the sine and cosine of the hue channel respectively.
The hue can be found as $y_{hue} =  \mathrm{atan2} \left(y_{shue}, y_{chue} \right)$.
We use the intermediate values when calculating the loss function.
During training we use the following loss function for each observation $y^\star$, and each corresponding prediction $y$.
\begin{align}
loss &= %
\frac{1}{2} \left(\sin(y^\star_{hue}) - y_{shue} \right)^2     \nonumber \\
&+ \frac{1}{2} \left(\cos(y^\star_{hue}) - y_{chue} \right)^2  \nonumber \\
&+ \left(y^\star_{sat} - y_{sat} \right)^2  \nonumber \\
&+ \left(y^\star_{val} - y_{val} \right)^2 %
\end{align}
The mean of this loss is taken over all observations in each mini-batch during training.
This loss function is continuous and correctly handles the wrap-around nature of the hue channel \parencite{WhiteRepresentingAnglesSE}.

\begin{figure}
	\newcommand{\picwidth}{60pt}
	\begin{tikzpicture}[defaultfig]
	
	\node (hiddenout)[layer, dotted]{Input Module};
	
	\node(outHue)[xshift=-1cm, above = 5 of hiddenout] {$y_{hue}$};
	\node(outSaturation)[above = 5 of hiddenout] {$y_{sat}$};
	\node(outValue)[xshift=1cm, above = 5 of hiddenout] {$y_{val}$};
	
	\node(Z)[above=-0.2cm of hiddenout.north]{};
	\node(Z2)[left=0cm of Z]{};
	\node(Z1)[left=0.3cm of Z2]{};
	\node(Z3)[right=0cm of Z]{};
	\node(Z4)[right=0.3cm of Z3]{};

	\node(s3)[layer, above=1 of Z3, xshift=0.5cm]{$\sigma$};
	\node(s4)[layer, above=1 of Z4, xshift=1cm]{$\sigma$};
	\draw[->](Z3) to (s3) to (outSaturation);
	\draw[->](Z4) to (s4) to (outValue);

	\node(AngHue)[layer, below = 1 of outHue, xshift=-0.7cm]{$ \mathrm{atan2} \left(y_{shue}, y_{chue} \right) $ };
	\draw[->](AngHue) to (outHue.south);
	
	\node(s1)[layer, above=1 of Z1, xshift=-1cm]{$\tanh$};
	\node(s2)[layer, above=1 of Z2, xshift=-0.5cm]{$\tanh$};
	\draw[->](Z1) to (s1);
	\draw[->](s1) to node[left]{$y_{shue}$} (AngHue);
	\draw[->](Z2) to (s2);
	\draw[->](s2) to node[right]{$y_{chue}$}  (AngHue.340);
	
	\end{tikzpicture}
	
	\caption{The Point Estimate Output Module.
		Here $\mathrm{atan2}$ is the quadrant preserving arctangent, outputting the angle in turns.
		\label{fig:pointoutmod}}
\end{figure}

\section{Evaluation}

\subsection{Perplexity in Color-Space}
Perplexity is a measure of how well the distribution, estimated by the model, matches the reality according to the observations in the test set.
Perplexity is commonly used for evaluating language models. Here however, it is being used to evaluate the discretized distribution estimate.
It can be loosely thought of as to how well the model's distribution does in terms of the size of an equivalent uniform distribution.
Note that this metric does not assume conditional independence of the color channels.

Here $\tau$ is the test-set made up of pairs consisting of a color name $t$, and a color-space point $\tilde{x}$;
and  $p(\tilde{x}\mid t)$ is the output of the evaluated model.
Perplexity is defined as:
\begin{equation}
PP(\tau) = \exp_2{\left(
	\frac{-1}{|\tau|} 
	\sum_{
		\forall(t,(\tilde{x})) \in \tau}
	\log_2 p(\tilde{x}\mid t)\right)}
\end{equation}

As the perplexity for a high-resolution discretized model will inherently be very large and difficult to read,
we define the standardized perplexity: $\frac{PP(\tau)}{n_{res}}$,
where $n_{res}$ is the total number of bins in the discretization scheme.
For all the results we present here $n_{res} = 256^3$.
This standardized perplexity gives the easily interpretable values \emph{usually} between zero and one.
It is equivalent to comparing the relative performance of the model to that of a uniform distribution of the same total resolution.
$\frac{PP(\tau)}{n_{res}}=1$ means that the result is equal to what we would see if we had distributed the probability mass uniformly into all bins in a 3D histogram.
$\frac{PP(\tau)}{n_{res}}=0.5$ means the result is twice as good as if we were to simply use a uniform distribution: it is equivalent to saying that the correct bin is selected as often as it would be had a uniform distribution with half as many bins been used (i.e. larger bins with twice the area).
The standardized perplexity is also invariant under different output resolutions.
Though for brevity we only  present results with 256 bins per channel, our preliminary results for using other resolutions are similar under standardized perplexity.


\subsection{Angularly Correct Calculations on HSV}\label{sec:angularly-correct}
We use the HSV color-space \parencite{smith1978color} throughout this work.
In this format: hue, saturation and value all range between zero and one.
Note that we measure hue in \emph{turns}, rather than the more traditional degrees, or radians.
Having hue measured between zero and one, like the other channels, makes the modeling task more consistent.
Were the hue to range between $0$ and $2\pi$ (radians) or between $0$ and $360$ (degrees) it would be over-weighted in the loss function and evaluation metrics compared to the other channels.
This regular space means that errors on all channels can be considered equally.
Unlike many other colors spaces (CIELab, Luv etc.) the gamut is square and all combinations of values from the different channels correspond to realizable colors.

When performing calculations with the HSV color-space, it is important to take into account that hue is an angle.
As we are working with the color-space regularized to range between zero and one for all channels,
this means that a hue of one and a hue of zero are equivalent (as we measure in turns, in radians this would be $0$ and $2\pi$).

The square error of two hue values is thus calculated as:
\begin{equation}
SE(h_1, h_2) = \min \big( \left(h_1 - h_2 \right)^2, \, \left(h_1 - h_2 -1 \right)^2  \big)
\end{equation}
This takes into account that the error can be calculated clockwise or counter-clockwise; and should be the smaller.
Note that the $-1$ term is related to using units of turns, were we using radians it would be $-2\pi$.

The mean of a set of hues ($\lbrace h_1,\,\ldots,\,h_N \rbrace$) is calculated as:
\begin{equation}
\bar h = \mathrm{atan2} \Bigg(%
	\frac{1}{N} \sum_{i=1}^{i=N} \sin (h_i), \,  %
	\frac{1}{N} \sum_{i=1}^{i=N} \cos (h_i)%
\Bigg)%
\end{equation}
This gives the mean angle.

\subsection{Non-compositional Baselines}
We consider a non-compositional model to establish a baseline on the color modeling part of this task; with the exclusion of the language understanding part.

The non-compositional methods do not process each term in the name; they do not work with the language at all.
They simply map from the exact input text (no tokenization) to the pre-calculated distribution or mean of the training data for the exact color name.
As there is plenty of training data for most color names (see \Cref{sec:fulltaskcorpusstats}) this is a very effective approach.
Strictly speaking, this \empmodel{} has less information than the neural network models
as it does not have the tokenized color name given to it, but only the whole name.
However, pragmatically learning to compose the sequence of terms into a meaningful whole is by far the harder part of this task.
This \empmodel{} bypasses the compositional language understanding part of the process.
It is as if the input module (as discussed in \Cref{sec:inputmod}) would perfectly resolve the sequence of terms into a single item.
These models can exploit the training observations without the need to determine how to compose the tokens.
This is a useful baseline, as our neural models (SOWE, CNN, GRU and LSTM)
each differs in how they compose the tokens, and on that this study focuses.

In theory the term-based neural models shoukd out-perform the \empmodel{}, if they learn a very good compositional understanding of the language.
This would require learning how the terms in the color name combine in a way that exceeds the information directly present in the training data per class.
\emph{It is this capacity of learning how the terms combine that allow for the models to predict the outputs for combinations of terms that never occur in the training data (\Cref{sec:extrapodata}).}
Learning a compositional model that exploits its term based knowledge in such a way  that generalizes to get better results than the direct exploitation of the training data (as in the \empmodel{}), is very difficult and would require very well calibrated control of (over/under)fitting.
This is particularly true in the case where there  is a  large amount of training data for the whole phrase.
Conversely, when there is no training data for the whole phrase (as considered in \Cref{sec:extrapodata}) non-compositional models can not function at all.

\subsubsection{Non-compositional Baseline for Distribution Estimation: KDE}\label{sec:direct-histogram} \label{sec:kernel-density-based-smoothing}
To define a \empmodel{} for the distribution estimation tasks,
we use kernel-density estimation (KDE) in a formulation for non-parametric estimation \parencite{silverman1986density} .
The KDE effectively produces a smoothed histogram from the training data as processed in \Cref{sec:discretization}.
It causes adjacent bins to have most similar probabilities, thus matching to the mathematical notion of a continuous random variable.
This is applied on-top of the histogram used for the training data.
We use the Fast Fourier Transform (FFT) based KDE method of the \textcite{silverman1982algorithm}.
We use a Gaussian kernel, and select the bandwidth per color description based on leave-one-out cross validation on the training data.
A known issue with the FFT-based KDE method is that it has a wrap-around effect near the boundaries, where the probability mass that would be assigned outside the boundaries is instead assigned to the bin on the other side.
For the value and saturation channels we follow the standard solution of initially defining additional bins outside the true boundaries, then discarding those bins and rescaling the probability to one.
For the hue channel this wrap-around effect is exactly as desired.

In our evaluations using KDE rather than just the training histograms directly proved much more successful on all distribution estimation tasks.
This is because it avoids empty bins, and effectually interpolates probabilities between observations.
We found in preliminary investigations that using KDE-based method to be much better than add-one smoothing.

We also investigated the application of KDE to the training data,  before training our term-based neural network based distribution models.
Results for this can be found in \Cref{sec:smoothed-results}.
In brief, we found that smoothing the training data does not significantly affect the result of the neural network based models.
As discussed in \Cref{sec:quantitative-results}, this is because the neural networks are able to learn the smoothness relationship of adjacent bins.

Our KDE-based \empmodel{} for distribution estimation bypasses the natural language understanding part of the task,
and directly uses the standard non-parametric probability estimation method to focus solely on modeling the distributions.
Matching its performance indicates that a model is effectively succeeding well at both the natural language understanding component and the distribution estimation component.

\subsubsection{Non-Compositional Baseline for Point Estimation: Mean-point}
In a similar approach, we also propose a method that directly produces a point estimate from a color name.
We define this by taking the mean (centroid) of all the training observations for a given exact color name.
The mean is taken in the angularly correct way (as discussed in \Cref{sec:angularly-correct}).
Taking the mean of all the observations gives the theoretically optimal solution to minimize the squared error on the training data set.
As with our direct distribution estimation method, this bypasses the term based language understanding, and directly exploits the training data.
It thus represents an approximate upper bound on the point estimation performance of the term based models.
Though, as discussed in \Cref{sec:intro}, the notion of mean and of minimizing the square error is not necessarily the correct way to characterize selecting the optimal point estimate for colors.
It is however a consistent way to do so, and so we use it for our evaluations.

\subsection{Evaluation Strategies and Data}

\subsubsection{Full Task}\label{sec:full-task}
We make use of the  Munroe dataset as prepared by \textcite{mcmahan2015bayesian} from the results of the XKCD color survey.
The XKCD color survey \parencite{Munroe2010XKCDdataset} collected over 3.4 million observations from over 222,500 respondents.
McMahan and Stone take a subset from Munroe's full survey, by restricting it to the responses from native English speakers, 
and removing very rare color names with less than 100 uses.
This gives a total of 2,176,417 observations and 829 color names. 
They also define a standard test, development and train split.

\paragraph{Full Task Corpus Statistics} \label{sec:fulltaskcorpusstats}
\begin{itemize}
	\item In the full corpus 829 unique color names made up of 308 unique terms. 
	
	\item Training split
	\begin{itemize}
		\item There are a total of 1,523,108 training observations.
		\item The distribution of observations between color names has the following quartile statistics: Q0: 70.0, Q1: 109.0, Q2: 214.0, Q3: 627.0, Q4: 152,953.0.
		\item The distribution of observations between terms has the following quartile statistics: Q0: 70.0, Q1: 148.5, Q2: 345.0, Q3: 2,241.75, Q4: 347,173.0.
	\end{itemize}

	\item Development split
	\begin{itemize}
		\item There are a total of 108,545 development observations.
		\item The distribution of observations between color names has the following quartile statistics:  Q0: 5.0, Q1: 7.0, Q2: 15.0, Q3: 45.0, Q4: 10,925.0.
		\item The distribution of observations between terms has the following quartile statistics: Q0: 5.0, Q1: 10.0, Q2: 24.5, Q3: 159.25, Q4: 24,754.0.
	\end{itemize}
	
	\item Test split
	\begin{itemize}
		\item There are a total of 544,764 testing observations.
		\item The distribution of observations between color names has the following quartile statistics:  Q0: 25.0, Q1: 40.0, Q2: 78.0, Q3: 225.0, Q4: 54,627.0.
		\item The distribution of observations between terms has the following quartile statistics: Q0: 26.0, Q1: 54.75, Q2: 124.5, Q3: 804.0, Q4: 124,138.0.
	\end{itemize}
\end{itemize}

\subsubsection{Unseen combination Task} \label{sec:extrapodata}
A primary interest in using the term based models is to be able to make predictions for never before seen descriptions of colors.
For example, based on the learned understanding of \texttt{salmon} and of \texttt{bright}, from examples like \texttt{bright green} and \texttt{bright red}, we wish for the system to make predictions about \texttt{bright salmon}, even though that description never occurs in the training data.
The ability to make predictions, such as these, illustrates term-based natural language understanding.
This cannot be done with the \empmodel{} models, which bypass the term processing step.
To evaluate this generalization capacity, we define new sub-datasets for both testing and training.
We select the rarest 100 color descriptions from the full dataset,
with the restriction that every token in a selected description must still have at least 8 uses in other descriptions in the training set.
The selected examples include multi-token descriptions such as: \texttt{bright yellow green} and also single tokens that occur more commonly as modifiers than as stand-alone descriptions such as \texttt{pale}.

The unseen combination testing set has only observations from the full test set that do use those rare descriptions.
We define a corresponding restricted training set made up of the data from the full training set, excluding those  corresponding to the rare descriptions.
A restricted development set is created similarly to the training set, containing data from the full (original) validation set, with the exclusion of rare descriptions used in the test set.
This was done so that no direct knowledge of the combined terms can leak during early-stopping.

By training on the restricted training set and testing on the unseen combinations, we can assess the model's capacity of compositionality to make predictions for color descriptions not seen during training.
A similar approach was used in \textcite{acl2018WinnLighter} and in \textcite{DBLP:journals/corr/AtzmonBKGC16}.
We contrast this to the same models when trained on the full training set to see how much accuracy was lost.

\paragraph{Unseen Combinations Corpus Statistics}
\begin{itemize}
	\item In the unseen combinations testset, there are (by design) 100 unique color names, that is 12.06\% of the full set of color names. Thus the number of unique color names in the restricted training set is decreased by 100 names (i.e 12.06\% smaller).
	\item 20,460 observations were removed from the training set . Thus the restricted training set contains 13.43\% fewer observations than the full training set.
	\item 14 terms are used across the 100 color names in the unseen combinations test set.
	They are \natlang{blue}, \natlang{bright}, \natlang{brown}, \natlang{dark}, \natlang{deep}, \natlang{dull}, \natlang{green}, \natlang{grey}, \natlang{ish}, \natlang{light}, \natlang{lime}, \natlang{olive}, \natlang{orange}, \natlang{pale}, \natlang{pink}, \natlang{purple}, \natlang{red}, \natlang{rose}, \natlang{teal}, \natlang{very}, \natlang{violet}, \natlang{y}, \natlang{yellow}, and \natlang{-}.
\end{itemize}

\subsubsection{Order Task}\label{sec:orderdata}
It is believed that the order of words in a color description matters, at least to some extent, for it's meaning.
For example, \natlang{greenish brown} and \natlang{brownish green} are distinct, if similar, colors.
To assess the models on their ability to make predictions when order matters we construct the order test set.
This is a subset of the full test set containing only descriptions with terms that occur in multiple different orders.
There are 76 such descriptions in the full dataset.
Each of which has exactly one alternate ordering.
This is unsurprising as while color descriptions may have more than 2 terms, normally one or more of the terms is a joining token such as \natlang{ish} or \natlang{-}.
We only construct an order testing set, and not a corresponding training set, as this is an evaluation using the model trained on the full training data.

\paragraph{Order Task Corpus Statistics}
\begin{itemize}
	\item 76 unique color names with 2 possible orders for their terms are used. They makes up 9.17\% of the unique color names in the full data set.
	\item In the full training set (which is used for training for this evaluation) there are 63,048 observations of these color names, making up 4.14\% of all training observations.
	\item 16 terms are used in these ambiguous ordered color names. Namely:
	\natlang{apple}, \natlang{blue}, \natlang{bright}, \natlang{brown}, \natlang{green}, \natlang{grey}, \natlang{ish}, \natlang{light}, \natlang{orange}, \natlang{pink}, \natlang{purple}, \natlang{red}, \natlang{violet}, \natlang{y}, \natlang{yellow} and \natlang{-}.
\end{itemize}

\section{Experimental Setup}

\subsection{Implementation}
The implementation of all the models was in the Julia programming language \parencite{Julia}.
The full implementation can be downloaded from the GitHub repository.\footnote{Implementation source is at \url{https://github.com/oxinabox/ColoringNames.jl}}
The machine learning components makes heavy use of the MLDataUtils.jl\footnote{MLDataUtils.jl is available from \url{https://github.com/JuliaML/MLDataUtils.jl}} and TensorFlow.jl \parencite{TensorFlowJulia}
 packages,
the latter of which was enhanced significantly to allow for this work to be carried out.
The discretization and the KDE for the \empmodel{} is done using KernalDensityEstimation.jl.%
\footnote{KernalDensityEstimation.jl  is available from \url{https://github.com/JuliaStats/KernelDensity.jl}}
The training data is managed with DataDeps.jl \parencite{2018arXiv180801091W}.

\subsection{Common Network Features}
Drop-out \parencite{srivastava2014dropout}  is used on all ReLU layers and on the recurrent units in the RNNs, with threshold of 0.5 during training.
The network is optimized using Adam \parencite{kingma2014adam}, and a learning rate of 0.001.
Early stopping is checked every 10 epochs using the development dataset.
Distribution estimation methods are trained using the full batch (where each observation is a distribution) for every epoch.
Point Estimation methods are trained using randomized mini-batches of $2^{16}$ observations (which are each color-space triples).
All hidden-layers, except as otherwise precluded (inside the convolution, and in the penultimate layer of the point estimation networks) have the same width 300, as does the embedding layer.

\section{Results}\label{sec:results}

\def\maincolors{%
	brown-orange,
	orange-brown,	
	yellow-orange,
	orange-yellow,
	brownish_green,
	greenish_brown,
	bluish_grey,
	greyish_blue,
	pink-purple,
	purple-pink,
	green,
	greenish,
	purple,	
	purplish,
	brown,
	brownish,
	black,
	grey,
	white%
}

\def\oovcolors{
	1Brown,
	1Green,
	1Purple,
	gray,
	1Gray%
}

\tikzset{lastline/.style={opacity=1}}
\newcommand{\distfigs}[2]{
	\begin{tikzpicture}[defaultfig]%
	\foreach[count=\mdlii from 0] \mdlfile/\mdlname in #1 {%
		\node at (3.1*\mdlii cm, -1cm) {\sffamily \mdlname};%
		\foreach[count=\colii from 0] \colorname in #2 {%
			\xdef\topy{1.1*\colii}
			\node at (3.1*\mdlii cm, \topy cm) %
			{\includegraphics[height=0.95cm]{%
					figs/demo/dist/\mdlfile/\colorname}%
			};%

		}%
		\xdef\topy{\topy+1}
		\node at (3.1*\mdlii cm, -1cm) {\sffamily \mdlname};%
		\node at (3.1*\mdlii cm, \topy) {\sffamily \mdlname};%
	}%
	%

	%
	\end{tikzpicture}%
}%

\subsection{Qualitative Results}\label{sec:qualitative-results}

\newcommand{\pointfigs}[3]{
	\begin{tikzpicture}%
	\foreach[count=\mdlii from 0] \mdlfile/\mdlname in #1 {%
		\node(fig\mdlfile) at (3.1*\mdlii cm, 0) %
		{\includegraphics[height=#3 cm]{%
				figs/demo/point/\mdlfile/#2}%
		};%
		\node[below = -0.7 of fig\mdlfile] {\sffamily \mdlname};%
		\node[above = -0.75 of fig\mdlfile] {\sffamily \mdlname};%
	}%
	\end{tikzpicture}%
}%

\begin{figure}
	\resizebox{\textwidth}{!}{
		\distfigs{{Direct/{Training Data}, Direct-smoothed/{\shortstack{Non-compositional\\ Baseline}}, SOWE/SOWE, CNN/CNN, GRU/GRU, LSTM/LSTM}}{\maincolors}
	}
	\caption{Some examples of the output distribution estimates from the models trained on the full dataset} \label{fig:distout1}
\end{figure}

\begin{figure}
	\resizebox{\textwidth}{!}{
		\pointfigs{{Direct/{\shortstack{Non-compositional\\ Baseline}}, SOWE/SOWE, CNN/CNN, GRU/GRU, LSTM/LSTM}}{maincolors}{18}
	}
	\caption{Some examples of the output point estimates from the models trained on the full dataset} \label{fig:pointout}
\end{figure}

To get an understanding of the problem and how the models are performing, we consider some of the outputs of the model for particular cases.
\Cref{fig:distout1} shows examples of distribution estimates, and 
\Cref{fig:pointout} shows similar examples for point estimates.
Both  are  taken  from models trained on  the full training dataset.
It can be seen that the models' outputs using term based estimation are generally similar to the non-term-based \empmodel{}, as is intended.
This shows that the models are correctly fitting to estimate the colors.
It can be noted  that  in general the  colors  are  very good, with  only a few marked exceptions,   discussed in the  following sections, particularly around multi-word colors.
To  the naked  eye,  it is hard to to distinguish  between the outputs of the different models.
The general high quality of the estimates aligns with the strong results found in the quantitative evaluations discussed in \Cref{sec:quantitative-results}.
The example shown  in \Cref{fig:distout1,fig:pointout} serve  to indicate that while the  quantitative results do show that some of the models perform better than  others, the  true visual difference is very small.

\subsubsection{On the effects of word-order}
The different input modules have a different capacity to leverage word-order.
This is reflected in \Cref{fig:distout1,fig:pointout},
when considering the pairs of outputs that differ only in word order, such as \natlang{purple-pink} and \natlang{pink-purple}.
The plots presented for the training data and for the \empmodel{} show that such color name pairs are subtly different but similar.
The SOWE model is unable to take into account word order at all, and so produces identical outputs for all orders.
The CNN models produce very similar outputs but not strictly identical -- spotting the difference requires a very close observation.
This is in-line with the different filter sizes allowing the CNN to effectively use n-gram features, and finding that the unigram features are the most useful.
Both RNN models (GRU and LSTM) produce estimated distributions that visibly depend on the order of words.
It seems that the first term dominates the final output: for example \natlang{greenish brown} is more green, and  \natlang{brownish green} is more brown, contrary to the linguistic understanding.
The RNN outputs are more similar to the color described by first term than any later terms.
We can see that the first term is not solely responsible for the final output however, as \natlang{purple-pink}, \natlang{purple} and \natlang{purplish} (tokenized as \natlang{purple}, \natlang{ish}) are all different.
It is surprising that the RNNs outputs  are dominated by the first term and not the latter terms\footnote{So much so that we double checked our implementation to be sure that it wasn't processing the inputs backwards.}.
This shows that they are  functioning to remember the earlier inputs.
However, they are struggling to attribute the significance of the word order.
Linguistically we would expect the last term to be the most significant:
\natlang{greenish brown} is a shade of brown, not green.
This expectation is reflected in the histogram for the training data.
Although, for many of the order swapped colors the training histograms shown are very similar regardless of the order.

\subsubsection{On the smoothness of the distribution estimates}\label{sec:learnedsmoothness}
In \Cref{fig:distout1} it can be seen that the term-based distribution estimation models are much smoother than the corresponding histograms taken from the  training data.
They are not as smooth as the \empmodel{} which explicitly uses KDE.
However, they are much smoother than would be expected, had the output bins been treated independently.
Thus it is clear that the models are learning that adjacent bins should have similar output values.
This is a common feature of all the training data, no matter which color is being described.
This learned effect is in line with the fact that color is continuous, and is only being represented here as discrete.
We note in relation to this learned smoothness: that while the models capture the highly asymmetrical shapes of most distributions well, they do not do well at capturing small dips.
Larger multi-modes as seen in the achromatic colors such as \natlang{white}, \natlang{grey}, \natlang{black}, \natlang{white}, are captured; but smaller dips such as the hue of \natlang{greenish} being more likely to be on either side of the green spectrum are largely filled in.
In general, it seems clear that additional smoothing of the training data is not required for the neural network based models.
This aligns with the results presented in \Cref{sec:smoothed-results}.


\pgfkeys{
	/pgf/number format/.cd, fixed, precision=3, fixed zerofill=true
}

\pgfplotstableset{
	col sep=comma,
	header=has colnames,
	column type={c},
	ignore chars={"},
	clear infinite,
	empty cells with={--},
	every head row/.style={before row=\toprule,	after row=\midrule},
	columns/method/.style={
		reset styles,
		string type,
		column name=Method,
		string replace*={Operational Upper Bound}{Non-compositional Baseline}},
	boldcell/.style = {
		postproc cell content/.append style={
			@cell content/.add={\mathversion{tabularbold}}{}
		}
	},
	greycell/.style = {
		postproc cell content/.append style={
			@cell content/.add={\color{gray}}{}
		}		
	}
}
\pgfplotstableset{distresults/.append style={%
		columns={method, perpstd},
		columns/perp/.style={column name=$PP$},
		create on use/perpstd/.style={
			create col/expr={\thisrow{perp}/(256*256*256)},
		},
		columns/perpstd/.style={column name=$\frac{PP}{256^3}$},
		every row 0 column 1/.style={greycell}
	},%
	extrapodistresults/.append style={
		every row 0 column 1/.style={greycell},
		every row 0 column 2/.style={greycell},
		columns={method, nxperpstd, xperpstd},
		columns/nxperpstd/.style={column name={\shortstack{\small Full\\Training Set\\$\frac{PP}{256^3}$}}},
		columns/xperpstd/.style={column name={\shortstack{\small Restricted\\Training Set\\$\frac{PP}{256^3}$}}},
		create on use/xperpstd/.style={
			create col/expr={\thisrow{extrapolatingperp}/(256*256*256)},
		},
		create on use/nxperpstd/.style={
			create col/expr={\thisrow{nonextrapolatingperp}/(256*256*256)},
		},
	}
}

\pgfplotstableset{pointresults/.append style={%
		columns={method,mse},
		columns/mse/.style={column name={$MSE$}},
		every row 0 column 1/.style={greycell},
		every row 5 column 1/.style={greycell},
	},%
	extrapopointresults/.append style={
		columns={method, nonextrapolatingmse, extrapolatingmse},
		columns/nonextrapolatingmse/.style={column name=\shortstack{\small Full\\Training Set\\$MSE$}},
		columns/extrapolatingmse/.style={column name=\shortstack{\small Restricted\\Training Set\\$MSE$}},
		every row 0 column 1/.style={greycell},
		every row 5 column 1/.style={greycell},
		every row 0 column 2/.style={greycell},
		every row 5 column 2/.style={greycell},
	},
}

\subsection{Quantitative Results}\label{sec:quantitative-results}

Overall, we see that our models are able to learn to estimate colors based on sequences of terms.
From the consideration of all the results shown in \Cref{tbl:pointfull,tbl:distfull,tbl:distord,tbl:pointord,tbl:pointextrapo,tbl:distextrapo}, 
the CNN and SOWE models perform almost as well as the \empmodel{}.
With the SOWE having a marginal lead for distribution estimation,
and the CNN and SOWE being nearly exactly equal for most point estimation tasks.
We believe the reason for this is that the SOWE is an easier to learn model from a gradient descent perspective: it is a shallow model with only one true hidden layer.
In general the results for the LSTM and GRU were very similar, and both much worse than the non-recurrent models.
While it is only marginally behind the SOWE and CNN on the full point estimation task (\Cref{tbl:pointfull}), on all other tasks for both point estimation and distribution estimation it is significantly worse.
This may indicate that it is hard to capture the significant relationships between terms in the sequence.
However, as discussed \Cref{sec:quantitative-results} it did learn generally acceptable colors to the human eye, but the quantitative results presented in this section show that it is not as close a match to the population's expectation.


\begin{table}
	\caption{\label{tbl:distfull} The results for the \textbf{full distribution estimation task}. Lower perplexity (PP) is better.}
	\begin{filecontents*}{results/regular/res_dist_full.csv}
"method","perp","mse_to_distmean"
"Operational Upper Bound",1.194494521492072e6,0.06636743143829865
"SOWE",1.2557197826221387e6,0.06794612557823515
"CNN",1.3080937965131702e6,0.06863270223255383
"GRU",1.4895938718144323e6,0.07664675511142244
"LSTM",1.5437051450855192e6,0.07704195445324569
\end{filecontents*}
\pgfplotstabletypeset[distresults,
	every row 1 column 1/.style={boldcell},
	]{results/regular/res_dist_full.csv}
\end{table}

\begin{table}
	\caption{\label{tbl:pointfull} The results for the \textbf{full point estimation task}. Lower mean squared error (MSE) is better.}
	\begin{filecontents*}{results/regular/res_point_comb_full.csv}
"method","mse"
"Operational Upper Bound",0.0663561224937439
"SOWE",0.06674081087112427
"CNN",0.06718235462903976
"GRU",0.07066445797681808
"LSTM",0.07121684402227402
"Distribution Mean Operational Upper Bound",0.06635611645652997
"Distribution Mean SOWE",0.06794612557823515
"Distribution Mean CNN",0.06863270223255383
"Distribution Mean GRU",0.07664675511142244
"Distribution Mean LSTM",0.07704195445324569
\end{filecontents*}
\pgfplotstabletypeset[pointresults,
	every row 2 column 1/.style={boldcell},
	every row 1 column 1/.style={boldcell}
	]{results/regular/res_point_comb_full.csv}
\end{table}

\subsubsection{Ordered Task}
The performance of SOWE on the order tasks (\Cref{tbl:distord,tbl:pointord}) is surprising.
For the distribution estimation it outperforms the CNN, and for point estimation it ties with the CNN.
The CNN and RNN, can take into account word order, but the SOWE model cannot.
The good results for SOWE suggest that the word-order is not very significant for color names.
While word order matters, different colors with the same terms in different order are similar enough for most colors that it still performs very well.
In theory the models that are capable of using word order have the capacity to ignore it, and thus could achieve a similar result.
An RNN can learn to perform a sum of its inputs (the word embeddings),
and the CNN can learn to weight all non-unigram filters to zero.
In practice we see that for the RNN in particular this clearly did not occur.
This can be attributed to the more complex networks being more challenging to train via gradient descent.
It seems that color-naming is not a task where word order substantially matters,
and thus the simpler SOWE model excels.


\begin{table}
	\caption{\label{tbl:distord} The results for the \textbf{order distribution estimation task}. Lower perplexity (PP) is better. This is a subset of the full test set containing only tests where the order of the words matters.}
	\begin{filecontents*}{results/regular/res_dist_ord.csv}
"method","perp","mse_to_distmean"
"Operational Upper Bound",893602.4850669863,0.06484226384736647
"SOWE",917009.8344842698,0.06560185760457346
"CNN",955151.1996613375,0.06582688105166774
"GRU",2.0837575219944436e6,0.09536452027977917
"LSTM",2.103330779848846e6,0.08765904141627993
\end{filecontents*}
\pgfplotstabletypeset[distresults,
	every row 1 column 1/.style={boldcell},
	]{results/regular/res_dist_ord.csv}
\end{table}

\begin{table}
	\caption{\label{tbl:pointord} The results for the \textbf{order point estimation task}. Lower mean squared error (MSE) is better. This is a subset of the full test set containing only tests where the order of the words matters.}
	\begin{filecontents*}{results/regular/res_point_comb_ord.csv}
"method","mse"
"Operational Upper Bound",0.06480130553245544
"SOWE",0.06605731695890427
"CNN",0.06559617072343826
"GRU",0.09559748321771622
"LSTM",0.09613459557294846
"Distribution Mean Operational Upper Bound",0.0648013100782032
"Distribution Mean SOWE",0.06560185760457346
"Distribution Mean CNN",0.06582688105166774
"Distribution Mean GRU",0.09536452027977917
"Distribution Mean LSTM",0.08765904141627993
\end{filecontents*}
\pgfplotstabletypeset[pointresults,
	every row 1 column 1/.style={boldcell},
	every row 2 column 1/.style={boldcell},
	%
	every row 5 column 1/.style={boldcell},
	every row 6 column 1/.style={boldcell}
	]{results/regular/res_point_comb_ord.csv}
\end{table}

\subsubsection{Unseen Combinations of Terms}
The SOWE and CNN models are able to generalize well to making estimates for combinations of color terms that are not seen in training.
\Cref{tbl:distextrapo,tbl:pointextrapo} show the results of the model on the test set made up of rare combinations of color names (as described in \Cref{sec:extrapodata}) for the restricted training set (which does not contain those terms).
These results on this test set are compared with the same models when trained on the full training set.
The \empmodel{} are unable to produce estimates from the unseen combinations testing set as they do not process the color names term-wise.
Performing well on this task is indicative as to if the models are learning how the terms combine to determine the color, as they cannot be simply matching the full color name (term sequence) against one that occurs in training.
This is an important test, as due to the combinatorial nature of language,
it is common to encounter term sequences in the real world that never occur during training.

On distribution estimation (\Cref{tbl:distextrapo}) the SOWE results are only marginally worse for the restricted training set as they are for the full training set.
The CNN results are worse again, but they are still better than the results on the full test-set.
The distribution estimates are good on absolute terms, having low evaluated perplexity.

In the point estimation task (\Cref{tbl:pointextrapo}) the order is flipped with the CNN outperforming the SOWE model.
In-fact the CNN actually performs better with the restricted training set for predicting the unseen test colors, than it does for predicting those colors when they are included in the full training set; though the difference is only marginal.
Unlike for distribution estimates, the unseen color point estimates are worse than the overall results from the full task (\Cref{tbl:pointfull}), though the errors are still small on an absolute scale.

Over all the performance of the SOWE and CNN remain strong on the unseen combination tasks.
The RNN models continue to perform poorly on the unseen combination of terms task for both point and distribution estimation.
The SOWE and CNN perform sufficiently well on the unseen combinations that the color estimates they produce would be practically useful.
The unseen combination results are comparable to the full dataset results discussed (shown in \Cref{tbl:distfull,tbl:pointfull}), and have very small errors on an absolute scale.


\begin{table}
	\caption{\label{tbl:distextrapo} The results for the \textbf{unseen combinations distribution estimation task}. Lower perplexity (PP) is better. This uses the unseen test set: a subset of the full test set contain only rare word combinations. In the restricted training set results these rare word combinations were removed from the training and development sets. In the full training set results the whole training and development stet was used, including the rare words that occur in the test set.}
	\begin{filecontents*}{results/regular/res_dist_extrapo.csv}
"method","nonextrapolatingperp","extrapolatingperp"
"Operational Upper Bound",843846.1755043188,NaN
"SOWE",831637.1322184883,929211.069918433
"CNN",867983.9518082727,1.0956519864598655e6
"GRU",1.9710975586917e6,3.0574071480120397e6
"LSTM",2.069032523950984e6,2.8895637598336423e6
\end{filecontents*}
\pgfplotstabletypeset[extrapodistresults,
	every row 1 column 1/.style={boldcell},
	every row 1 column 2/.style={boldcell}
	]{results/regular/res_dist_extrapo.csv}
\end{table}

\begin{table}
	\caption{\label{tbl:pointextrapo} The results for the \textbf{unseen combinations point estimation task}. Lower mean squared error (MSE) is better. This uses the unseen test set: a subset of the full test set contain only rare word combinations. In the restricted training set results these rare word combinations were removed from the training and development sets. In the full training set results the whole training and development stet was used, including the rare words that occur in the test set.}
	\begin{filecontents*}{results/regular/res_point_comb_extrapo.csv}
"method","nonextrapolatingmse","extrapolatingmse"
"Operational Upper Bound",0.06213238462805748,NaN
"SOWE",0.06493569165468216,0.07881485670804977
"CNN",0.07246111333370209,0.06999575346708298
"GRU",0.13774675130844116,0.14229418337345123
"LSTM",0.13809597492218018,0.14087577164173126
"Distribution Mean Operational Upper Bound",0.06213238168891163,NaN
"Distribution Mean SOWE",0.07285586328364832,0.07639481398062851
"Distribution Mean CNN",0.07291856105657271,0.08437807415514435
"Distribution Mean GRU",0.10478952562356844,0.15215027128488232
"Distribution Mean LSTM",0.10499460876378672,0.11197592739284144
\end{filecontents*}
\pgfplotstabletypeset[extrapopointresults,
	every row 1 column 1/.style={boldcell},
	every row 2 column 2/.style={boldcell},
	]{results/regular/res_point_comb_extrapo.csv}
\end{table}

\subsubsection{Extracting the mean from the distribution estimates}

In the point estimation results discussed so far have been from models trained specifically for point estimation (as described by \Cref{sec:point-estimation}).
However, it is also possible to derive the mean from the distribution estimation models.
Those results are also presented in \Cref{tbl:pointfull,tbl:pointord,tbl:pointextrapo}.
In general these results perform marginally worse (using the MSE metric) than their corresponding modules using the point estimation output module.
The only exception to this is the LSTM for both the unseen combination tasks and the order task, for which it was notably better to use the mean from the distribution rather than one directly trained.
We note that for the \empmodel{}, the distributions mean is almost identical to the true mean of points, as expected.

\subsubsection{On the differences between the distribution estimation and point estimation training procedure}

Beyond the output module there are a few key differences between the point estimation modules and the distribution estimate modules.
When training distribution estimation models, all examples of a particular color name is grouped into a single high information training observation using the histogram as the output.
Whereas when training for point estimation, each example is processed individually (using minibatches).
This means that the distribution estimating models fit to all color names with equal priority.
Whereas for point estimates, more frequently used color names have more examples, and so more frequent color names are fit with priority over rarer ones.
Another consequence of using training per example using random minibatches, rather than aggregating and training with full batch, is increased resilience to to local minima \parencite{lecun2012efficient}.
One of the upsides of the aggregated training used in distribution estimation is that it trains much faster as only a small number of high-information training examples are processed, rather than a much larger number of individual observations.

It may be interesting in future work to consider training the distribution estimates per example using one-hot output representations; thus making the process similar to that used in the point estimate training.
It is  possible that such a method may have trouble learning the smoothness of the output space (as discussed in \Cref{sec:learnedsmoothness}),
as it would not see demonstration of the partial activation of adjacent bins in the training examples.
However, this is not certain,  much like the point estimation trained on one-hot learns a representation that minimises mean squared error outputting  a point  between  all the training examples, it is reasonable to  expect that the distribution estimates will output a smooth histogram  as  this is near to a minimum for the cross-entropy.
With the current model the presence of partial activation of adjacent bins in all examples  may be causing the smoothness to  be learned primarily in the output layer, and with little respect for the inputs.
Such would explain the difficulties in capturing subtler features of the output distribution, such as the depth of the valley between the two peaks in the hue of \natlang{greenish} shown in \Cref{fig:distout1}.
Using one-hot examples for training, may help force encoding the knowledge of the nature of continuous distributions deeper into the network allowing the input color name to have a more pronounce effect.

\subsection{Training set results}

To investigate our supposition that the SOWE, is a much easier function to fit via gradient descent, as compared to the CNN or the RNNs, we consider the error rate on the full training set during the training of the models.
These plots are shown in \Cref{fig:disttrainloss} and \Cref{fig:pointtrainloss}.
These plots seem to support the supposition, as the SOWE training error decreases notably faster (it is a steeper curve) in both cases.
This corresponds to a easier error surface in network parameter (weights and biases) space, with fewer points of low gradient, or near local minima.
If we compare the final loss of each method on the training set (before it was stopped due to early stopping) against the test set results in  \Cref{tbl:distfull,tbl:pointfull}
we find they are similar,  particularly for distribution estimation ( \Cref{tbl:distfull} and \Cref{fig:disttrainloss}).
While for point estimation (\Cref{fig:pointtrainloss,tbl:pointfull}), on the test set CNN and SOWE perform similarly, while RNNs perform much worse, despite the fact that in training the performance of CNN is roughly midway between SOWE and the RNNs.
In all cases, the absolute error in training has become small relative the to the natural variation in the training set by the time early stopping terminates training.
Note that perfect fit is not possible as the training data varies.


\pgfplotsset{
	trainingloss/.style={
		cycle list name=exotic,
		xlabel = {Epoch},
		mark = x,
 		y tick label style={
			/pgf/number format/.cd,
			fixed,
			fixed zerofill,
			precision=3,
			/tikz/.cd
		},
		x tick label style={
			/pgf/number format/.cd,
			fixed,
			precision=0,
			/tikz/.cd
		},
	},
}

\begin{figure}
	\begin{minipage}{0.7\textwidth}
		\begin{tikzpicture}
			\begin{axis}[
			trainingloss,
			title={Distribution Estimators, Training Error},
			ylabel={$\frac{PP}{256^3}$},
			]
			\addplot table [y=SOWE-STDPERP, x=Step, col sep=comma] {results/training/distest-perplexity.csv};
			\addplot table [y=CNN-STDPERP, x=Step, col sep=comma] {results/training/distest-perplexity.csv};
			\addplot table [y=GRU-STDPERP, x=Step, col sep=comma] {results/training/distest-perplexity.csv};
			\addplot table [y=LSTM-STDPERP, x=Step, col sep=comma] {results/training/distest-perplexity.csv};
			\legend{{SOWE},{CNN},{GRU},{LSTM}};
			\end{axis}
		\end{tikzpicture}
	\end{minipage}
	\hspace{0.5cm}
	\begin{minipage}{0.2\textwidth}
	Final Training Performance\\
	\pgfplotstabletypeset[columns/res/.style={column name=$\frac{PP}{256^3}$}]{
		method, res
		SOWE, 0.0571749749549454
		CNN, 0.0699227578631911
		GRU, 0.080777487788597
		LSTM, 0.0937867587201507
	}
	\end{minipage}
	\caption{The training set error of the distribution estimation models, when trained on the full dataset. Note that the plots stop when the model ceased training due to the development set error rising (early stopping). \label{fig:disttrainloss}}.
\end{figure}

\begin{figure}
	\begin{minipage}{0.7\textwidth}
		\begin{tikzpicture}
		\begin{axis}[
			trainingloss,
			title={Point Estimators, Training Error},
			ylabel={MSE},
			scaled y ticks={real:3}, 
			ytick scale label code/.code={},
			]
			\addplot table [y=SOWE, x=Step, col sep=comma] {results/training/pointest.csv};
			\addplot table [y=CNN, x=Step, col sep=comma] {results/training/pointest.csv};
			\addplot table [y=GRU, x=Step, col sep=comma] {results/training/pointest.csv};
			\addplot table [y=LSTM, x=Step, col sep=comma] {results/training/pointest.csv};
			\legend{{SOWE},{CNN},{GRU},{LSTM}}
		\end{axis}
		\end{tikzpicture}
	\end{minipage}
	\hspace{0.5cm}
	\begin{minipage}{0.2\textwidth}
		Final Training Performance\\
	\pgfplotstabletypeset{
		method, MSE
		SOWE, 0.13222543895244598	
		CNN, 0.13548657298088074
		GRU, 0.14894478023052216
		LSTM, 0.14776922762393951
	}
	\end{minipage}

	\caption{The training set error of the point estimation models, when trained on the full dataset. Note that the plots stop when the model ceased training due to the development set error rising (early stopping). \label{fig:pointtrainloss}}
\end{figure}

\subsection{Completely Unseen Color Estimation From Embeddings}\label{sec:embeddingonly}
As an interesting demonstration of how the models function by learning the transformation from the embedding space to the output, we briefly consider the outputs for color-names that do not occur in the training or testing data at all.
This is even more extreme than the unseen combination task considered in \Cref{tbl:distextrapo,tbl:pointextrapo} where the terms appeared in training, but not the combination of terms.
In the examples shown in \Cref{fig:oovdist,fig:oovpoint}, where the terms never occurred in the training data at all, our models exploit the fact that they work by transforming the word-embedding space to predict the colors.
There is no equivalent for this in the direct models.
While \natlang{Grey} and \natlang{gray} never occur in the training data; \natlang{grey} does, and it is near-by in the word-embedding space.
Similar is true for the other colors that vary by capitalization.
We only present a few examples of single term colors here, and no quantitative investigation, as this is merely a matter of interest.

It is particularly interesting to note that the all the models make similar estimations for each color.
This occurs both for point estimation and for distribution estimation.
They do well on the same colors and make similar mistakes on the colors they do poorly at.
The saturation of \natlang{Gray} is estimated too high, making it appear too blue/purple, this is also true of \natlang{grey} though to a much lesser extent.
\natlang{Purple} and \natlang{Green} produce generally reasonable estimates.
The hue for \natlang{Brown} is estimated as having too much variance, allowing the color to swing into the red or yellowish-green parts of the spectrum.
This suggests that in general all models are learning a more generally similar transformation of the space.
In general the overall quality of each model seems to be in line with that found in the results for the full tests.

\begin{figure}
	\distfigs{{SOWE/SOWE, CNN/CNN, GRU/GRU, LSTM/LSTM}}{\oovcolors}	
	\caption{Some example distribution estimations for colors names which are completely outside the training data. The terms: \natlang{Brown}, \natlang{gray}, \natlang{Gray}, \natlang{Green}, and \natlang{Purple}, do not occur in any of the color data; however \natlang{brown}, \natlang{grey} \natlang{green}, and \natlang{purple} do occur.} \label{fig:oovdist}
\end{figure}

\begin{figure}
	\pointfigs{{SOWE/SOWE, CNN/CNN, GRU/GRU, LSTM/LSTM}}{oovcolors}{5}
	\caption{Some example point estimates for colors names which are completely outside the training data. The terms: \natlang{Brown}, \natlang{gray}, \natlang{Gray}, \natlang{Green}, and \natlang{Purple}, do not occur in any of the color data; however \natlang{brown}, \natlang{grey} \natlang{green}, and \natlang{purple} do occur.} \label{fig:oovpoint}
\end{figure}

\section{Conclusion}
We have presented four input modules (SOWE, CNN, GRU and LSTM),
and two output modules (distribution estimate, and point estimate)
that are suitable for using machine learning to make estimates about color based on the terms making up its name.
We contrasted these to a \empmodel{} model for each task which bypassed the term-wise natural language understanding component of the problem.
We found the results for SOWE, and CNN were strong, approaching this strong baseline.

It is a note-worthy feature on the current state of short phrase modeling,
and the difficulty of compositional natural language understanding
that the term-based models are not able to out-perform the \empmodel{} where training data for the whole phrases was available.
The term-based models are effectively given additional information, in the form of the tokenisation, but are unable to fully leverage it in the general case.
They are unable to outperform simply ignoring the common sub-phase information when training instances for the whole phrase are available.

A key take away from our results is that using a SOWE should be preferred over an RNN for short phrase natural language understanding tasks when order is not a very significant factor.
It is also important to evaluate if order is indeed a significant factor, since on the surface one would expect it to be for color names.
One way to evaluate this is to include SOWE as a baseline model in other tasks.
While RNNs are the standard type of model for problems with sequential input, such as color names made up of multiple words as we considered here.
However, we find both LSTM and GRU performance to be significantly exceeded by SOWE and CNN.
SOWE is an unordered model roughly corresponding to a bag of words.
CNN similarly roughly correspondents to a bag of ngrams, in our case a bag of all 1,2,3,4 and 5-grams.
This means that the CNN can readily take advantage of both fully ordered information, using the filters of length 5, down to unordered information using  filters of length 1.
RNNs however must fully process the ordered nature of its inputs, as its output comes only from the final node.
Between the two RNN models it seems the GRU performs marginally better.
It would be interesting to further compare with bidirectional variants of these RNNs.

In a broader context, we envisage the distribution learned for a color name can be used as a prior probability, and when combining with additional context information, a likelihood can be estimated for particular uses.
This additional information could take the from of other words, such as estimating the distribution for a \natlang{brown dog}, as compared to a \natlang{brown tree}, or from other sources.
A particularly interesting related avenue for investigation would condition the model not only on the words used but also on the speaker.
The original source of the data, \textcite{Munroe2010XKCDdataset}, includes some demographic information which is not explored as a model input in any  published model (to the best of our knowledge).
It is expected that color-term usage may vary with subcultures.

\clearpage
\bibliography{master}

\begin{thebibliography}{42}
\expandafter\ifx\csname natexlab\endcsname\relax\def\natexlab#1{#1}\fi

\bibitem[{Atzmon et~al.(2016)Atzmon, Berant, Kezami, Globerson, and
  Chechik}]{DBLP:journals/corr/AtzmonBKGC16}
Atzmon, Yuval, Jonathan Berant, Vahid Kezami, Amir Globerson, and Gal Chechik.
  2016.
\newblock Learning to generalize to new compositions in image understanding.
\newblock \emph{CoRR}, abs/1608.07639.

\bibitem[{Berk, Kaufman, and Brownston(1982)}]{Berk:1982:HFS:358589.358606}
Berk, Toby, Arie Kaufman, and Lee Brownston. 1982.
\newblock A human factors study of color notation systems for computer
  graphics.
\newblock \emph{Commun. ACM}, 25(8):547--550.

\bibitem[{Berlin and Kay(1969)}]{berlin1969basic}
Berlin, Brent and Paul Kay. 1969.
\newblock \emph{Basic color terms: Their university and evolution}.
\newblock California UP.

\bibitem[{Bezanson et~al.(2014)Bezanson, Edelman, Karpinski, and Shah}]{Julia}
Bezanson, Jeff, Alan Edelman, Stefan Karpinski, and Viral~B. Shah. 2014.
\newblock {J}ulia: A fresh approach to numerical computing.
\newblock \emph{SIAM Review}, 59(1):65--98.

\bibitem[{Bojanowski et~al.(2017)Bojanowski, Grave, Joulin, and
  Mikolov}]{bojanowski2016enriching}
Bojanowski, Piotr, Edouard Grave, Armand Joulin, and Tomas Mikolov. 2017.
\newblock Enriching word vectors with subword information.
\newblock \emph{Transactions of the Association for Computational Linguistics},
  5:135--146.

\bibitem[{Cho et~al.(2014)Cho, van Merri{\"e}nboer, Bahdanau, and
  Bengio}]{cho2014properties}
Cho, Kyunghyun, Bart van Merri{\"e}nboer, Dzmitry Bahdanau, and Yoshua Bengio.
  2014.
\newblock On the properties of neural machine translation: Encoder-decoder
  approaches.
\newblock In \emph{Eighth Workshop on Syntax, Semantics and Structure in
  Statistical Translation (SSST-8)}.

\bibitem[{Conway(1992)}]{conway1992experimental}
Conway, Damian. 1992.
\newblock An experimental comparison of three natural language colour naming
  models.
\newblock In \emph{Proc. east-west int. conf. on human-computer interaction},
  pages 328--339.

\bibitem[{Gers, Schmidhuber, and Cummins(1999)}]{gers1999learning}
Gers, Felix~A, J{\"u}rgen Schmidhuber, and Fred Cummins. 1999.
\newblock Learning to forget: Continual prediction with lstm.

\bibitem[{Heider(1972)}]{heider1972universals}
Heider, Eleanor~R. 1972.
\newblock Universals in color naming and memory.
\newblock \emph{Journal of experimental psychology}, 93(1):10.

\bibitem[{Heider and Olivier(1972)}]{HEIDER1972337}
Heider, Eleanor~Rosch and Donald~C. Olivier. 1972.
\newblock The structure of the color space in naming and memory for two
  languages.
\newblock \emph{Cognitive Psychology}, 3(2):337 -- 354.

\bibitem[{Hochreiter and Schmidhuber(1997)}]{hochreiter1997long}
Hochreiter, Sepp and J{\"u}rgen Schmidhuber. 1997.
\newblock Long short-term memory.
\newblock \emph{Neural computation}, 9(8):1735--1780.

\bibitem[{Jones and Lotwick(1984)}]{jones1984remark}
Jones, MC and HW~Lotwick. 1984.
\newblock Remark as r50: a remark on algorithm as 176. kernal density
  estimation using the fast fourier transform.
\newblock \emph{Journal of the Royal Statistical Society. Series C (Applied
  Statistics)}, 33(1):120--122.

\bibitem[{Kawakami et~al.(2016)Kawakami, Dyer, Routledge, and
  Smith}]{DBLP:journals/corr/KawakamiDRS16}
Kawakami, Kazuya, Chris Dyer, Bryan~R. Routledge, and Noah~A. Smith. 2016.
\newblock Character sequence models for colorfulwords.
\newblock \emph{CoRR}, abs/1609.08777.

\bibitem[{Kelly et~al.(1955)}]{kelly1955iscc}
Kelly, Kenneth~Low et~al. 1955.
\newblock Iscc-nbs method of designating colors and a dictionary of color
  names.

\bibitem[{Kingma and Ba(2014)}]{kingma2014adam}
Kingma, Diederik and Jimmy Ba. 2014.
\newblock Adam: A method for stochastic optimization.
\newblock \emph{arXiv preprint arXiv:1412.6980}.

\bibitem[{Lammens(1994)}]{ele1994computational}
Lammens, Johan Maurice~Gisele. 1994.
\newblock \emph{A Computational Model of Color Perception and Color Naming}.
\newblock Ph.D. thesis, State University of New York.

\bibitem[{LeCun et~al.(2012)LeCun, Bottou, Orr, and
  M{\"u}ller}]{lecun2012efficient}
LeCun, Yann~A, L{\'e}on Bottou, Genevieve~B Orr, and Klaus-Robert M{\"u}ller.
  2012.
\newblock Efficient backprop.
\newblock In \emph{Neural networks: Tricks of the trade}. Springer, pages
  9--48.

\bibitem[{Leshno et~al.(1993)Leshno, Lin, Pinkus, and Schocken}]{leshno1993uat}
Leshno, Moshe, Vladimir~Ya Lin, Allan Pinkus, and Shimon Schocken. 1993.
\newblock Multilayer feedforward networks with a nonpolynomial activation
  function can approximate any function.
\newblock \emph{Neural networks}, 6(6):861--867.

\bibitem[{Malmaud and White(2018)}]{TensorFlowJulia}
Malmaud, Jonathan and Lyndon White. 2018.
\newblock Tensorflow.jl: An idiomatic julia front end for tensorflow.
\newblock \emph{Journal of Open Source Software}.

\bibitem[{{Mansimov} et~al.(2015){Mansimov}, {Parisotto}, {Lei Ba}, and
  {Salakhutdinov}}]{2015arXiv151102793M}
{Mansimov}, E., E.~{Parisotto}, J.~{Lei Ba}, and R.~{Salakhutdinov}. 2015.
\newblock {Generating Images from Captions with Attention}.
\newblock \emph{ArXiv e-prints}.

\bibitem[{McMahan and Stone(2015)}]{mcmahan2015bayesian}
McMahan, Brian and Matthew Stone. 2015.
\newblock A bayesian model of grounded color semantics.
\newblock \emph{Transactions of the Association for Computational Linguistics},
  3:103--115.

\bibitem[{Menegaz et~al.(2007)Menegaz, Le~Troter, Sequeira, and
  Boi}]{menegaz2007discrete}
Menegaz, Gloria, Arnaud Le~Troter, Jean Sequeira, and Jean-Marc Boi. 2007.
\newblock A discrete model for color naming.
\newblock \emph{EURASIP Journal on Applied Signal Processing},
  2007(1):113--113.

\bibitem[{Meo, McMahan, and Stone(2014)}]{meomcmahanstone:color}
Meo, T., B.~McMahan, and M.~Stone. 2014.
\newblock Generating and resolving vague color reference.
\newblock \emph{Proc. 18th Workshop Semantics and Pragmatics of Dialogue
  (SemDial)}.

\bibitem[{Mojsilovic(2005)}]{mojsilovic2005computational}
Mojsilovic, Aleksandra. 2005.
\newblock A computational model for color naming and describing color
  composition of images.
\newblock \emph{IEEE Transactions on Image Processing}, 14(5):690--699.

\bibitem[{{Monroe}, {Goodman}, and {Potts}(2016)}]{2016arXiv160603821M}
{Monroe}, W., N.~D. {Goodman}, and C.~{Potts}. 2016.
\newblock {Learning to Generate Compositional Color Descriptions}.
\newblock \emph{ArXiv e-prints}.

\bibitem[{Monroe et~al.(2017)Monroe, Hawkins, Goodman, and
  Potts}]{DBLP:journals/corr/MonroeHGP17}
Monroe, Will, Robert X.~D. Hawkins, Noah~D. Goodman, and Christopher Potts.
  2017.
\newblock Colors in context: {A} pragmatic neural model for grounded language
  understanding.
\newblock \emph{CoRR}, abs/1703.10186.

\bibitem[{Munroe(2010)}]{Munroe2010XKCDdataset}
Munroe, Randall. 2010.
\newblock Xkcd: Color survey results.

\bibitem[{Mylonas et~al.(2015)Mylonas, Purver, Sadrzadeh, MacDonald, and
  Griffin}]{mylonas2015use}
Mylonas, Dimitris, Matthew Purver, Mehrnoosh Sadrzadeh, Lindsay MacDonald, and
  Lewis Griffin. 2015.
\newblock The use of english colour terms in big data.
\newblock The Color Science Association of Japan.

\bibitem[{van~den Oord et~al.(2016)van~den Oord, Dieleman, Zen, Simonyan,
  Vinyals, Graves, Kalchbrenner, Senior, and
  Kavukcuoglu}]{DBLP:journals/corr/OordDZSVGKSK16}
van~den Oord, A{\"{a}}ron, Sander Dieleman, Heiga Zen, Karen Simonyan, Oriol
  Vinyals, Alex Graves, Nal Kalchbrenner, Andrew~W. Senior, and Koray
  Kavukcuoglu. 2016.
\newblock Wavenet: {A} generative model for raw audio.
\newblock \emph{CoRR}, abs/1609.03499.

\bibitem[{Oord, Kalchbrenner, and Kavukcuoglu(2016)}]{oord2016pixel}
Oord, Aaron van~den, Nal Kalchbrenner, and Koray Kavukcuoglu. 2016.
\newblock Pixel recurrent neural networks.
\newblock \emph{arXiv preprint arXiv:1601.06759}.

\bibitem[{Reed et~al.(2016)Reed, Akata, Yan, Logeswaran, Schiele, and
  Lee}]{reed2016generative}
Reed, Scott, Zeynep Akata, Xinchen Yan, Lajanugen Logeswaran, Bernt Schiele,
  and Honglak Lee. 2016.
\newblock Generative adversarial text to image synthesis.
\newblock In \emph{Proceedings of The 33rd International Conference on Machine
  Learning}, volume~3.

\bibitem[{Silverman(1982)}]{silverman1982algorithm}
Silverman, BW. 1982.
\newblock Algorithm as 176: Kernel density estimation using the fast fourier
  transform.
\newblock \emph{Journal of the Royal Statistical Society. Series C (Applied
  Statistics)}, 31(1):93--99.

\bibitem[{Silverman(1986)}]{silverman1986density}
Silverman, B.W. 1986.
\newblock \emph{Density Estimation for Statistics and Data Analysis}.
\newblock Chapman \& Hall/CRC Monographs on Statistics \& Applied Probability.
  Taylor \& Francis.

\bibitem[{Smith(1978)}]{smith1978color}
Smith, Alvy~Ray. 1978.
\newblock Color gamut transform pairs.
\newblock \emph{ACM Siggraph Computer Graphics}, 12(3):12--19.

\bibitem[{Sonoda and Murata(2017)}]{SONODA2017uat}
Sonoda, Sho and Noboru Murata. 2017.
\newblock Neural network with unbounded activation functions is universal
  approximator.
\newblock \emph{Applied and Computational Harmonic Analysis}, 43(2):233 -- 268.

\bibitem[{Srivastava et~al.(2014)Srivastava, Hinton, Krizhevsky, Sutskever, and
  Salakhutdinov}]{srivastava2014dropout}
Srivastava, Nitish, Geoffrey Hinton, Alex Krizhevsky, Ilya Sutskever, and
  Ruslan Salakhutdinov. 2014.
\newblock Dropout: A simple way to prevent neural networks from overfitting.
\newblock \emph{The Journal of Machine Learning Research}, 15(1):1929--1958.

\bibitem[{Van De~Weijer et~al.(2009)Van De~Weijer, Schmid, Verbeek, and
  Larlus}]{van2009learning}
Van De~Weijer, Joost, Cordelia Schmid, Jakob Verbeek, and Diane Larlus. 2009.
\newblock Learning color names for real-world applications.
\newblock \emph{IEEE Transactions on Image Processing}, 18(7):1512--1523.

\bibitem[{{White} et~al.(2018){White}, {Togneri}, {Liu}, and
  {Bennamoun}}]{2018arXiv180801091W}
{White}, L., R.~{Togneri}, W.~{Liu}, and M.~{Bennamoun}. 2018.
\newblock {DataDeps.jl: Repeatable Data Setup for Replicable Data Science}.
\newblock \emph{ArXiv e-prints}.

\bibitem[{White(2016)}]{WhiteRepresentingAnglesSE}
White, Lyndon. 2016.
\newblock Encoding angle data for neural networks.
\newblock Cross Validated Stack Exchange.

\bibitem[{White et~al.(2015)White, Togneri, Liu, and
  Bennamoun}]{White2015SentVecMeaning}
White, Lyndon, Roberto Togneri, Wei Liu, and Mohammed Bennamoun. 2015.
\newblock How well sentence embeddings capture meaning.
\newblock In \emph{Proceedings of the 20th Australasian Document Computing
  Symposium}, ADCS '15, pages 9:1--9:8, ACM.

\bibitem[{White et~al.(2018)White, Togneri, Liu, and
  Bennamoun}]{novelperspective}
White, Lyndon, Roberto Togneri, Wei Liu, and Mohammed Bennamoun. 2018.
\newblock Novelperspective: Identifying point of view characters.
\newblock In \emph{Proceedings of ACL 2018, System Demonstrations}, Association
  for Computational Linguistics.

\bibitem[{Winn and Muresan(2018)}]{acl2018WinnLighter}
Winn, Olivia and Smaranda Muresan. 2018.
\newblock 'lighter' can still be dark: Modeling comparative color descriptions.
\newblock In \emph{Proceedings of the 56th Annual Meeting of the Association
  for Computational Linguistics (Volume 2: Short Papers)}, pages 790--795,
  Association for Computational Linguistics.

\end{thebibliography}

\clearpage
\appendix

\section{Appendix}

\subsection{On the Conditional Independence of Color Channels given a Color Name}\label{sec:corrind}

As discussed in the main text, we conducted a superficial investigation into the truth of our assumption that given a color name, the distributions of the hue, value and saturation are statistically independent.

We note that this investigation is, by no means, conclusive though it is suggestive.
The investigation focusses around the use of the Spearman's rank correlation.
This correlation measures the monotonicity of the relationship between the random variables.
A key limitation is that the relationship may exist but be non-monotonic.
This is almost certainly true for any relationship involving channels, such as hue, which wrap around.
In the case of such relationships Spearman's correlation will underestimate the true strength of the relationship.
Thus, this test is of limited use in proving conditional independence.
However, it is a quick test to perform and does suggest that the conditional independence assumption may not be so incorrect as one might assume.

In Monroe Color Dataset the training data  given by $V \subset \mathbb{R}^{3}\times T$, where $\mathbb{R}^{3}$ is the value in the color-space under consideration, and $T$ is the natural language space.
The subset of the training data for the description $t \in T$ is given by
$V_{|t}=\{(\tilde{v}_i,\,t_i) \in V \: \mid \: t_{i}=t\}$.
Further let $T_V = \{t_i \: \mid \: (\tilde{v},t_i)\in V$ be the set of color names used in the training set.
Let $V_{\alpha|t}$ be the $\alpha$ channel component of $V_{|t}$, i.e. $V_{\alpha|t} = \left\lbrace v_\alpha \mid ((v_1,v_2,v_3), t) \in V_{|t} \right\rbrace$.

The set of absolute Spearman's rank correlations between channels $a$ and $b$ for each color name is given by
$S_{ab}=\left\lbrace \left|\rho(V_{a|t},V_{b|t})\,\right|\,t\in T_{V}\right\rbrace$.
\newpage

We consider the third quartile of that correlation as the indicative statistic in \Cref{tbl:colorcor}.
That is to say for 75\% of all color names, for the given color-space, the correlation is less than this value.

Of the 16 color-spaces considered, it can be seen that the HSV exhibits the strongest signs of conditional independence -- under this (mildly flawed) metric.
More properly put, it exhibits the weakest signs of non-independence.
This includes being significantly less correlated than other spaces featuring circular channels such as HSL and HSI.

Our overall work makes the conditional independence assumption, much like n-gram language models make the Markov assumption.
The success of the main work indicates that the assumption does not cause substantial issues.

\pgfkeys{/pgf/number format/.cd, fixed relative, precision=4}
\pgfplotstableset{corstyle/.append style={%
		col sep=tab,
		header=has colnames,
		columns/Color Space/.style={reset styles, string type, column name=Color-Space},
		ignore chars={"},
		every head row/.style={before row=\toprule,	after row=\midrule}
	}%
}

\begin{table}
	\caption{\label{tbl:colorcor} The third quartile for the pairwise Spearman's correlation of the color channels given the color name.}
	\centering
		\begin{filecontents*}{results/colorcor.tsv}
Color Space	$Q3(S_{12})$	$Q3(S_{13})$	$Q3(S_{23})$	max
HSV	0.186133418362448	0.186659440736361	0.162773544394482	0.186659440736361
HSL	0.165517311460995	0.214707486317326	0.311289469985504	0.311289469985504
YCbCr	0.400460518342826	0.439294990723562	0.337702155482984	0.439294990723562
YIQ	0.408830959370809	0.49752505716774	0.406433708334577	0.49752505716774
LCHab	0.525834465160308	0.411030235162374	0.368758257040075	0.525834465160308
DIN99d	0.544230769230769	0.442607300988028	0.480269371641465	0.544230769230769
DIN99	0.544947433518862	0.493092269212655	0.523501840349666	0.544947433518862
DIN99o	0.560819648895458	0.408223817910852	0.521137044026191	0.560819648895458
RGB	0.603046817316173	0.447192698064549	0.565590231875486	0.603046817316173
Luv	0.559793543871927	0.611177475128859	0.437940012818417	0.611177475128859
LCHuv	0.612351195718699	0.407194033946441	0.341580575113509	0.612351195718699
HSI	0.244608678821158	0.239117831051628	0.630169457923765	0.630169457923765
CIELab	0.573018053092617	0.459749888953429	0.639028975560264	0.639028975560264
xyY	0.723026037679802	0.502412033968836	0.416540706605223	0.723026037679802
LMS	0.968036710423174	0.745753424657534	0.778987828315182	0.968036710423174
XYZ	0.972601133476646	0.816650275382821	0.784359038656462	0.972601133476646
\end{filecontents*}
\pgfplotstabletypeset[corstyle]{results/colorcor.tsv}
\end{table}

\subsection{KDE based smoothing of Training Data}\label{sec:smoothed-results}

It can be seen that smoothing has very little effect on the performance of any of the neural network based distribution estimation models.
All four term based models (SOWE, CNN, LSTM, GRU all perform very similarly whether or not the training data is smoothed.
This is seen consistently in all the distribution estimation tasks.
Contrast \Cref{tbl:distfull-smoothed,tbl:distord-smoothed,tbl:distextrapo-smoothed}
to the tables for the unsmoothed results
\Cref{tbl:distfull,tbl:distord,tbl:distextrapo}.

If however, smoothing is not applied to the operational upper bound, it works far worse.
In  \Cref{tbl:distfull-smoothed,tbl:distord-smoothed,tbl:distextrapo-smoothed} the Direct result refers to using the training histograms almost directly, without any smoothing or term-based input processing.
This is the same as the operational upper bound, minus the KDE.
It works very poorly (by comparison).
This is because the bins values are largely independent: a very high probability in one bin does not affect the probability of the adjacent bin -- which by chance of sampling may be lower than would be given by the true distribution.

This is particularly notable in the case of the direct, full training set result on the unseen combinations task reported  in \Cref{tbl:distextrapo-smoothed}. As these were some of the rarest terms in the training set, several did not coincide with any bins for observations in testing set.
This is because without smoothing it results in estimating the probability based on bins unfilled by any observation.
We do cap that empty bin probability at $\epsilon_{64} \approxeq 2\times 10^{-16}$ to prevent undefined perplexity.
We found capping the lower probability for bins like this to be far more effective than add-on smoothing.

Conversely, on this dataset the neural network models do quite well, with or without smoothing.
As the network can effectively learn the smoothness, not just from the observations of one color but from all of the observations.
It learns that increasing the value of one bin should increase the adjacent ones.
As such smoothing does not need to be applied to the training data.

\pgfkeys{
	/pgf/number format/.cd, fixed, precision=3, fixed zerofill=true
}

\pgfplotstableset{
	col sep=comma,
	header=has colnames,
	column type={c},
	ignore chars={"},
	string replace={Direct}{\emph{Direct}},
	clear infinite,
	empty cells with={--},
	every head row/.style={before row=\toprule,	after row=\midrule},
	columns/method/.style={reset styles, string type, column name=Method},
	boldcell/.style = {
		postproc cell content/.append style={
			@cell content/.add={\mathversion{tabularbold}}{}
		}
	},
	greycell/.style = {
		postproc cell content/.append style={
			@cell content/.add={\color{gray}}{}
		}		
	}
}
\pgfplotstableset{distresults/.append style={%
		columns={method, perpstd},
		columns/perp/.style={column name=$PP$},
		create on use/perpstd/.style={
			create col/expr={\thisrow{perp}/(256*256*256)},
		},
		columns/perpstd/.style={column name=$\frac{PP}{256^3}$},
		every row 0 column 1/.style={greycell}
	},%
	extrapodistresults/.append style={
		every row 0 column 1/.style={greycell},
		every row 0 column 2/.style={greycell},
		columns={method, nxperpstd, xperpstd},
		columns/nxperpstd/.style={column name={\shortstack{\small Full\\Training Set\\$\frac{PP}{256^3}$}}},
		columns/xperpstd/.style={column name={\shortstack{\small Restricted\\Training Set\\$\frac{PP}{256^3}$}}},
		create on use/xperpstd/.style={
			create col/expr={\thisrow{extrapolatingperp}/(256*256*256)},
		},
		create on use/nxperpstd/.style={
			create col/expr={\thisrow{nonextrapolatingperp}/(256*256*256)},
		},
	}
}

\begin{table}
	\caption{\label{tbl:distfull-smoothed}  The results for the \textbf{full distribution estimation task} using smoothed training data. Lower perplexity (PP) is better. This corresponds to the main results in \Cref{tbl:distfull}.}
	\begin{filecontents*}{results/smoothed/res_dist_full.csv}
"method","perp","mse_to_distmean"
"Direct",2.753563202939139e6,0.06635611645652997
"Operational Upper Bound",1.194494521492072e6,0.06636743143829865
"SOWE-smoothed",1.2652720569271764e6,0.0680232175890777
"CNN-smoothed",1.3222704773117197e6,0.06882467782095193
"GRU-smoothed",1.4693672333856223e6,0.07728473596794
"LSTM-smoothed",1.5091774666100473e6,0.07563614455197812
\end{filecontents*}
\pgfplotstabletypeset[distresults,
	every row 1 column 1/.style={greycell},
	]{results/smoothed/res_dist_full.csv}
\end{table}

\begin{table}
	\caption{\label{tbl:distord-smoothed}  The results for the \textbf{order distribution estimation task} using smoothed training data. Lower perplexity (PP) is better. This is a subset of the full test set containing only tests where the order of the words matters. This corresponds to the main results in \Cref{tbl:distord}.}
	\begin{filecontents*}{results/smoothed/res_dist_ord.csv}
"method","perp","mse_to_distmean"
"Direct",4.0885180011922754e6,0.0648013100782032
"Operational Upper Bound",893602.4850669863,0.06484226384736647
"SOWE-smoothed",916068.7637769555,0.06571741897544901
"CNN-smoothed",979343.738728208,0.0658741117229477
"GRU-smoothed",2.045028940870233e6,0.09662407256631958
"LSTM-smoothed",2.0077494650982902e6,0.0886307536054804
\end{filecontents*}
\pgfplotstabletypeset[distresults,
	every row 1 column 1/.style={greycell},
	]{results/smoothed/res_dist_ord.csv}
\end{table}

\begin{table}
	\caption{\label{tbl:distextrapo-smoothed}  The results for the \textbf{unseen combinations distribution estimation task} using smoothed training data. Lower perplexity (PP) is better. This uses the extrapolation subset of the full test set. In the extrapolating results certain rare word combinations were removed from the training and development sets. In the non-extrapolating results the full training and development stet was used. This corresponds to the main results in \Cref{tbl:distextrapo}.}
	\begin{filecontents*}{results/smoothed/res_dist_extrapo.csv}
"method","nonextrapolatingperp","extrapolatingperp"
"Direct",2.950834193953115e9,NaN
"Operational Upper Bound",843846.1755043188,NaN
"SOWE-smoothed",839005.141480351,942045.6940659512
"CNN-smoothed",892993.5811751804,1.0548725620390475e6
"GRU-smoothed",1.8800725169582975e6,3.0749299428323586e6
"LSTM-smoothed",2.0012037772128677e6,2.7254877905014763e6
\end{filecontents*}
\pgfplotstabletypeset[extrapodistresults,
	every row 1 column 1/.style={greycell},
	every row 1 column 2/.style={greycell},
	]{results/smoothed/res_dist_extrapo.csv}
\end{table}

\end{document}